\documentclass[10pt,twocolumn,letterpaper]{article}

\usepackage{cvpr}              % To produce the CAMERA-READY version

\usepackage[accsupp]{axessibility}  % Improves PDF readability for those with disabilities.

\usepackage{graphicx}
\usepackage{amsmath}
\usepackage{amssymb}
\usepackage{booktabs}
\usepackage{multirow}

\usepackage{amsthm}
\usepackage{caption}
\usepackage{subcaption}

\usepackage{algorithm}
\usepackage{algpseudocode}
\usepackage{amsmath,amssymb,amsfonts}
\usepackage{colortbl}

\newcommand{\vct}[1]{\ensuremath{\mathbf{#1}}}

\usepackage{accents}
\usepackage{amsmath,amssymb,amsfonts}
\usepackage{nicefrac}      
\usepackage{algorithm}
\usepackage{algpseudocode}
\usepackage{gensymb}

\usepackage{bm}
\def\mW{{\bm{W}}}

\usepackage[pagebackref,breaklinks,colorlinks]{hyperref}

\usepackage[capitalize]{cleveref}
\crefname{section}{Sec.}{Secs.}
\Crefname{section}{Section}{Sections}
\Crefname{table}{Table}{Tables}
\crefname{table}{Tab.}{Tabs.}

\def\cvprPaperID{7360} % *** Enter the CVPR Paper ID here
\def\confName{CVPR}
\def\confYear{2022}
\definecolor{Gray}{gray}{0.92}

\begin{document}

\title{Improving the Transferability of Targeted Adversarial Examples through Object-Based Diverse Input} %Realistic 3D Model
\author{Junyoung Byun\quad Seungju Cho\quad Myung-Joon Kwon\quad Hee-Seon Kim\quad Changick Kim\\
Korea Advanced Institute of Science and Technology (KAIST)\\
{\tt\small \{bjyoung, joyga, kwon19, hskim98, changick\}@kaist.ac.kr}
}
\maketitle

%%%%%%%%% ABSTRACT
\begin{abstract}
The transferability of adversarial examples allows the deception on black-box models, and transfer-based targeted attacks have attracted a lot of interest due to their practical applicability. To maximize the transfer success rate, adversarial examples should avoid overfitting to the source model, and image augmentation is one of the primary approaches for this. However, prior works utilize simple image transformations such as resizing, which limits input diversity. To tackle this limitation, we propose the object-based diverse input (ODI) method that draws an adversarial image on a 3D object and induces the rendered image to be classified as the target class. Our motivation comes from the humans' superior perception of an image printed on a 3D object. If the image is clear enough, humans can recognize the image content in a variety of viewing conditions. Likewise, if an adversarial example looks like the target class to the model, the model should also classify the rendered image of the 3D object as the target class. The ODI method effectively diversifies the input by leveraging an ensemble of multiple source objects and randomizing viewing conditions. In our experimental results on the ImageNet-Compatible dataset, this method boosts the average targeted attack success rate from 28.3\% to 47.0\% compared to the state-of-the-art methods. We also demonstrate the applicability of the ODI method to adversarial examples on the face verification task and its superior performance improvement. Our code is available at \url{https://github.com/dreamflake/ODI}.
\end{abstract} % against unknown models
\vspace{-0.1cm}
\captionsetup[sub]{font=small}

%%%%%%%%% BODY TEXT
\section{Introduction}
\label{sec:intro}
Deep learning models have demonstrated outstanding performance in a variety of fields and have permeated our daily lives \cite{he2016deep,deng2019arcface,esteva2019guide}. However, adversarial examples show that these models are vulnerable to maliciously crafted small input perturbations \cite{szegedy2014intriguing, dong2018boosting}. Interestingly, an adversarial example that is generated to attack a network is likely to disturb other networks as well. This intriguing property of adversarial examples is known as \textit{transferability} \cite{papernot2016transferability,li2020towards,wang2021enhancing}. This property allows an adversary to attack a black-box target model without knowing its interior.

\begin{figure}[t]
     \centering
         \centering
         \includegraphics[width=0.47\textwidth,trim={0cm 0.5cm 0cm 0cm},clip]{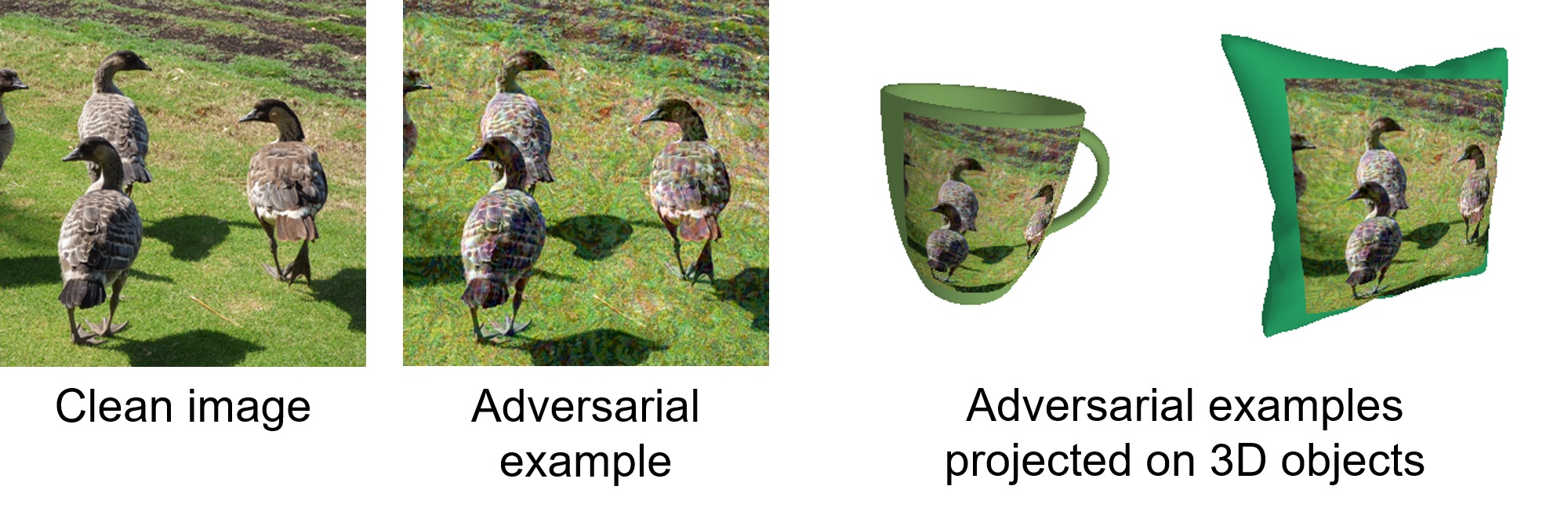}
         \vspace{-0.2cm}
         \caption{
        \textbf{Illustrations of our motivation.} If a targeted adversarial example really looks like the target class to the model, the model should also classify the adversarial examples projected on the 3D objects as the target class.}
        \label{fig:fig1}
        \vspace{-0.2cm}
\end{figure}
\begin{figure}[t]
     \centering
         \includegraphics[width=0.475\textwidth,trim={0cm 0cm 0cm 0cm},clip]{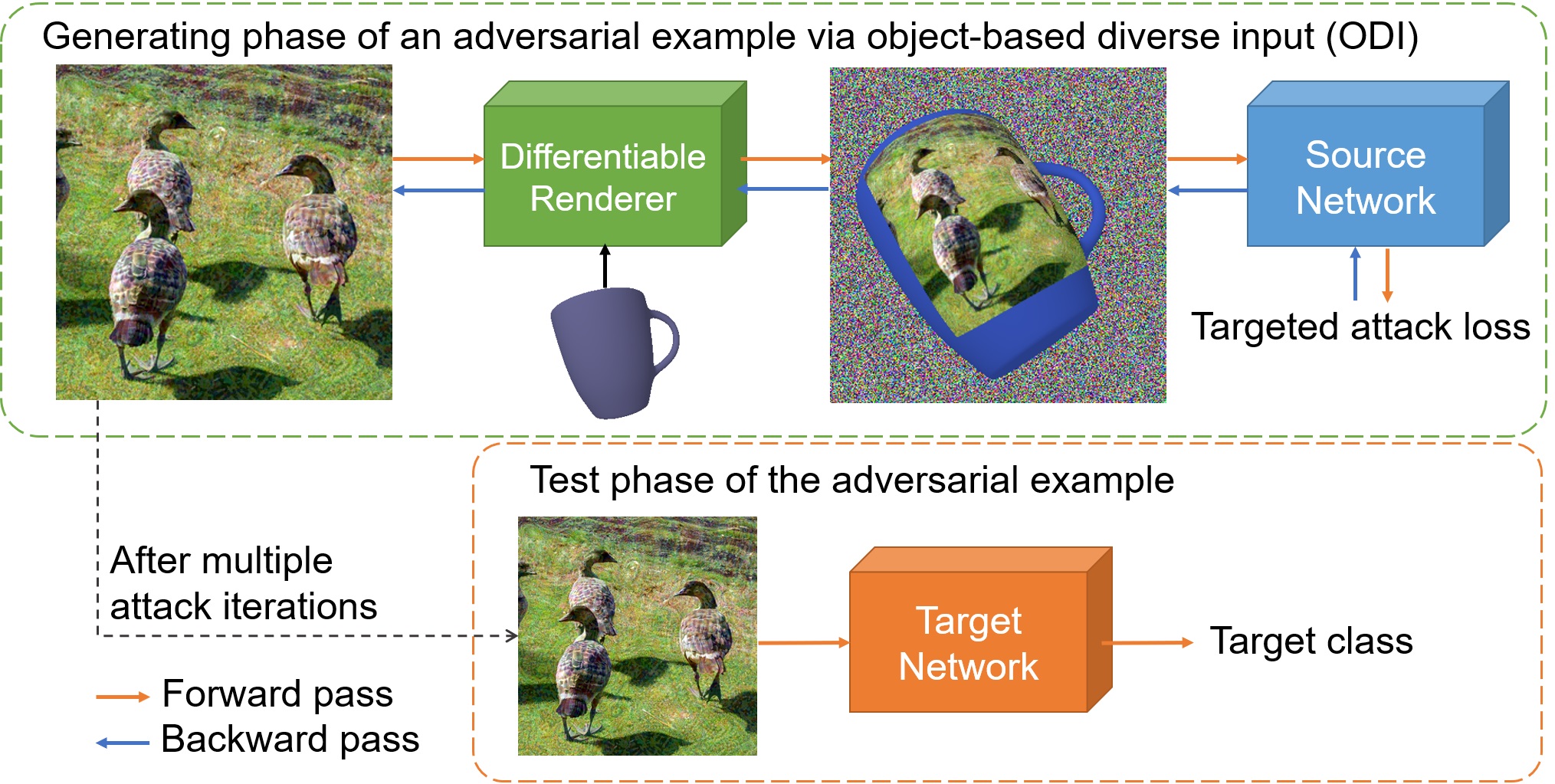}
        \caption{\textbf{The framework of targeted adversarial attacks with the proposed object-based diverse input (ODI) method.} Please note that the ODI method exploits 3D adversarial objects in the generating phase only. It finally improves the transferability of 2D adversarial examples.}
        \label{fig:fig2}
        \vspace{-0.5cm}
\end{figure}
% without  a target class
On black-box models, targeted attacks are significantly more challenging compared to non-targeted attacks which simply induce the victim models to malfunction without specifying a  target class \cite{li2020towards, zhao2020success}. Targeted attacks demand further exploration since they can cause more serious problems by deceiving models into predicting a designated harmful target class. Research on these transfer-based targeted attacks is important since it can help service providers prepare their models against these potential threats and assess the robustness of their models.

The transfer success rates greatly vary depending on the difference between the source and target models. Various approaches have been presented to improve the transferability, such as introducing momentum \cite{dong2018boosting,lin2019nesterov} and different loss functions \cite{li2020towards,zhao2020success} for better optimization, input data augmentation \cite{xie2019improving,zou2020improving}, and utilizing an ensemble of multiple source models \cite{liu2016delving}.

Among these strategies, we focus on input transformation-based methods and their limitations. These methods create adversarial examples that are robust against image transforms such as random resizing \cite{xie2019improving,zou2020improving} and translation \cite{dong2019evading} to prevent overfitting to the source model. However, since these methods use simple image transformations, they limit the diversity of input.

Our motivation for tackling this limitation comes from the humans' superior perception of an image printed on a 3D object (\eg, promotional merchandise commonly distributed at event booths). As a 2D image is projected on a 3D object, the original image is bent, the color looks different due to illumination, and some parts of the image are invisible depending on the viewpoint. Nevertheless, if the image is clear enough, humans can recognize the image content on the 3D object in a variety of viewing conditions. Likewise, if an adversarial example really looks like the target class to the source model, the model should also recognize the target class in the image printed on 3D objects. Our motivation is illustrated in~\cref{fig:fig1}.

From this motivation, we propose the \textbf{object-based  diverse input (ODI)} method for boosting the transferability of targeted adversarial examples. Specifically, we introduce 3D objects and project an adversarial example on the objects' surfaces. Then, we induce the rendered objects to be classified as the target class in a variety of rendering environments, including different lighting and viewpoints. This realistic input diversification can generalize the attack ability and improve the transferability of the adversarial example. The overall scheme is illustrated in~ \cref{fig:fig2}.

Our contributions can be listed as follows.
\begin{itemize} 
    \item We propose the object-based diverse input  (ODI) method to enhance the transferability of targeted adversarial examples. To our knowledge, this is the first time that 3D objects are used as canvases for 2D adversarial examples during their optimizations.
    \item We discovered that the attack success rate varies depending on the 3D object (\eg, a pillow and a cup). Our experimental results also indicate that an ensemble of carefully chosen source objects can further improve  transferability.
    \item In the experimental results with four source models and ten target models on the ImageNet dataset, the proposed ODI method boosts the average targeted attack success rate from 28.3\% to 47.0\% compared to the combination of state-of-the-art methods. 
    \item We also demonstrate the applicability of the ODI method to adversarial examples on the face verification task and its overwhelming performance improvement.
\end{itemize}

\begin{figure*}[t]
     \centering
     \begin{subfigure}[b]{0.97\textwidth}
         \centering
         \includegraphics[width=\textwidth,trim={0cm 0cm 0cm 0cm},clip]{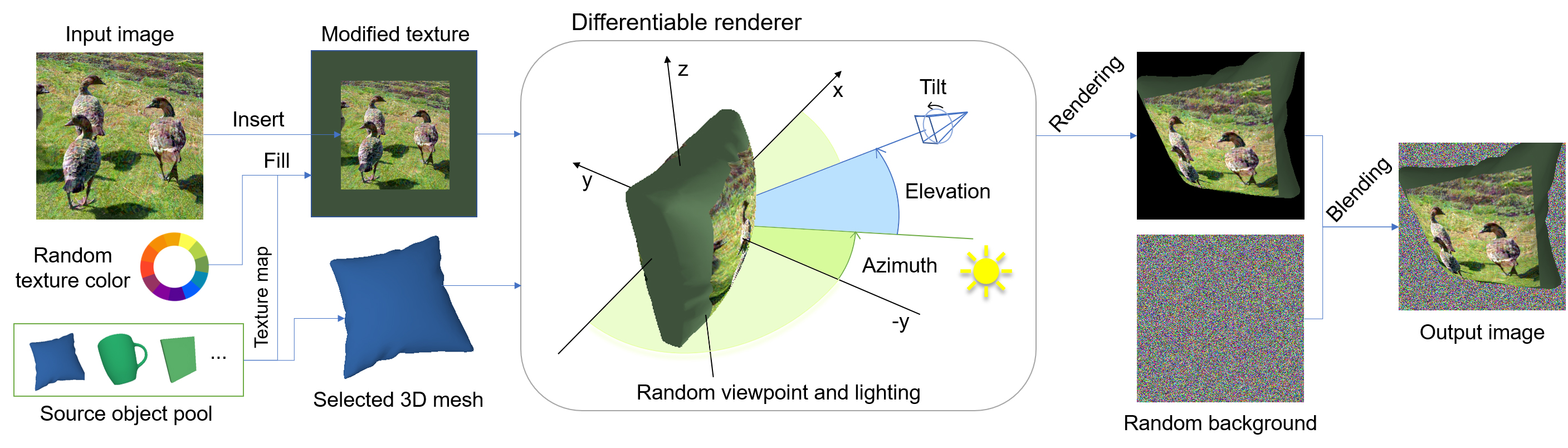}
     \end{subfigure}
        \vspace{-0.2cm}
        \caption{The pipeline of the object-based diverse input (ODI) method.}
        \label{fig:fig3}
        \vspace{-0.3cm}
\end{figure*}

\section{Related Work}
\label{sec:related_work}
\subsection{Adversarial Attacks in the Black-Box Setting}

In the black-box setting, since adversaries cannot access the interiors of the target model, the gradient of the image cannot be directly calculated using backpropagation. Query-based attacks \cite{brendel2018decision,chen2017zoo,cheng2019sign} use multiple queries to find an adversarial example via gradient estimation \cite{chen2017zoo, cheng2019sign} or random search \cite{brendel2018decision,ACFH2020square}. However, they are based on the unreasonable assumption that the output of the target model can be obtained via queries.

In comparison, transfer-based attacks \cite{li2020towards, wu2020boosting,lu2020enhancing} can generate adversarial examples that deceive the target model without requiring a query. Specifically, transfer-based attack methods generate adversarial examples through a white-box attack on a surrogate model, and the attacker expects the transferability of the adversarial example and attempts to deceive the target model with the generated image \cite{papernot2016transferability,papernot2017practical}. Therefore, we need to generate highly transferable adversarial examples that are capable of deceiving unknown models using a white-box surrogate model. 

Adversarial attacks in the white-box scenario take advantage of the gradient of the loss function with respect to the image. The standard algorithm for $\ell_\infty$-norm-constrained adversarial perturbations utilizes the sign of the gradient, which is called the fast gradient sign method (FGSM) \cite{goodfellow2014explaining}.
Formally, let $f$ be the classifier and $\mathcal{L}$ be the loss function for targeted attacks. Then, the targeted adversarial example $\vct x_{adv}$ can be found by solving the following optimization problem, given an image $\vct x$ and a target label $y_{t}$.
\begin{equation}
    % x_{adv} = x - \alpha \cdot \text{sign}(\frac{\partial L(x,y_{target})}{\partial x})
    \vct x_{adv} = \vct x - \epsilon \cdot \text{sign}(\nabla_{\vct x} \mathcal{L}(f(\vct x),y_{t})),
    \label{eqn:adv}
\end{equation}
where $\epsilon$ represents the step size. It updates the image $\vct x$ to minimize the loss for targeted attacks. It can be further optimized by iterating updates $\vct x$ on~\cref{eqn:adv} with a smaller step size $\alpha$, and this iterative version is called iterative-FGSM (I-FGSM) \cite{kurakin2016adversarial}. To aid in avoiding local minima to improve transferability, Dong \etal \cite{dong2018boosting} incorporate a momentum term in the optimization, which is referred to as momentum iterative FGSM (MI-FGSM).

In addition to these fundamental adversarial attacks, various techniques have been proposed to improve transferability by helping the image avoid falling into local minima and prevent overfitting to a specific source model.

One common approach is input diversification. The diverse inputs (DI) method \cite{xie2019improving} applies random resizing and padding to the image with probability $p$ for each inference in the iterative attacks to minimize overfitting to the source model. The resized-diverse-inputs (RDI) method \cite{zou2020improving} is similar to the DI method, but it resizes the expanded and padded image back to its original size at the final step of the DI. Unlike DI, which stochastically applies the image transform, RDI always applies the resizing image transform (\ie, $p=1$). 

% TI
The translation-invariant (TI) attack method \cite{dong2019evading} computes a weighted average of gradients from a set of translated images within a specified range, giving a higher weight to smaller displacements. To minimize computing time, TI approximates the weighted mean of the gradients by applying Gaussian blur to the original gradient. Updating the image with the obtained gradient mitigates the adversarial example's overfitting to the source model.

% ATTA
Wu \etal \cite{wu2021improving} highlight the limitations of heuristic image transformations like resizing \cite{xie2019improving} and propose the adversarial transformation-enhanced transfer attack (ATTA). They train an adversarial transformation network that neutralizes adversarial examples within an adversarial learning framework. Then, they construct a more robust adversarial example that is resistant to the trained adversarial transform. However, since their adversarial transformation network is based on a 2-layer CNN, the network can perform only simple image transformations, such as blurring and sharpening.

%SI 
The scale-invariant (SI) attack method \cite{lin2019nesterov} generates several scale variants of an image by altering the scale of pixel values and computing the gradient from them for each iteration. This promotes the transferability of adversarial examples by minimizing overfitting to the source model.

% VT 
The recently proposed variance tuning (VT) method \cite{wang2021enhancing} focuses on gradient variance, defined as the difference between an image's gradient and the average gradients of nearby images. By reducing the gradient variance, this can stabilize the update direction. % at the prior step

On the other hand, various approaches have been presented to improve transferability by using different loss functions for targeted attacks. Li \etal \cite{li2020towards}
identify the issue of the widely used cross-entropy loss, which results in vanishing gradient problems in iterative targeted attacks. To address this issue and increase the transferability, they adapt the size of the gradients using the Poincare distance. Zhao \etal \cite{zhao2020success} point out that prior work uses an inappropriately small number of iterations in the optimization of targeted adversarial examples. They emphasize that the following simple logit loss $\mathcal{L}_{logit}$ for targeted attacks can achieve state-of-the-art performance with sufficient iterations.
\begin{equation}
    \mathcal{L}_{logit}( f(\vct x^{adv}),y_t)=-\ell_t(f(\vct x^{adv})),
    \label{eqn:logit}
\end{equation}
where $\ell_t$ is the logit output corresponding to the target class.

\subsection{Adversarial Attacks with Differentiable Rendering}
Differentiable rendering projects 3D objects onto 2D images, and by making the internal process differentiable, it allows computing the gradient of the 3D objects' attributes, such as mesh and texture \cite{kato2020differentiable}. This differentiable rendering enables the optimization of 3D objects via digital simulation, and is commonly used to generate physically applicable adversarial examples that are robust under various viewpoints \cite{athalye2018synthesizing,zeng2019adversarial,xiao2019meshadv}. The most significant difference between these methods and the ODI method is that the prior works treat an adversarial mesh as their \textbf{goal}, but the ODI method treats it as a \textbf{tool} for improving the transferability of a 2D adversarial image.

\begin{figure}[t]

     \centering
\resizebox{0.4\textwidth}{!}{%
     \begin{subfigure}[b]{0.155\textwidth}
         \centering
         \includegraphics[width=\textwidth,trim={0cm 0cm 0cm 0cm},clip]{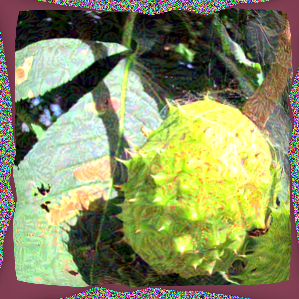}
         \caption{Default view}
     \end{subfigure}
          \begin{subfigure}[b]{0.155\textwidth}
         \centering
         \includegraphics[width=\textwidth,trim={0cm 0cm 0cm 0cm},clip]{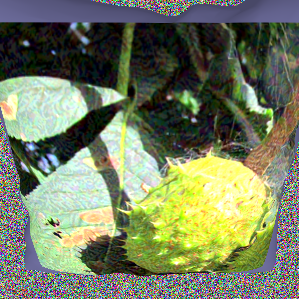}
         \caption{Elevation $+25\degree$}
     \end{subfigure}
          \begin{subfigure}[b]{0.155\textwidth}
         \centering
         \includegraphics[width=\textwidth,trim={0cm 0cm 0cm 0cm},clip]{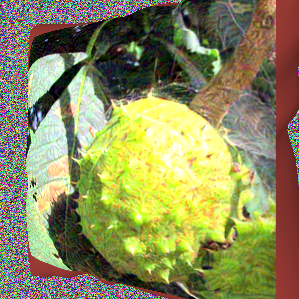}
         \caption{Azimuth $+25\degree$}
     \end{subfigure}
        }
        \resizebox{0.4\textwidth}{!}{%
 \begin{subfigure}[b]{0.155\textwidth}
         \centering
         \includegraphics[width=\textwidth,trim={0cm 0cm 0cm 0cm},clip]{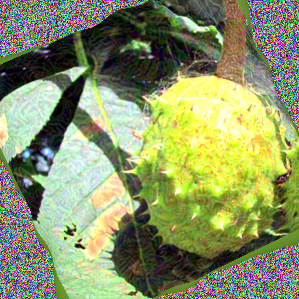}
         \caption{Tilt $+25\degree$}
     \end{subfigure}
         \begin{subfigure}[b]{0.155\textwidth}
         \centering
         \includegraphics[width=\textwidth,trim={0cm 0cm 0cm 0cm},clip]{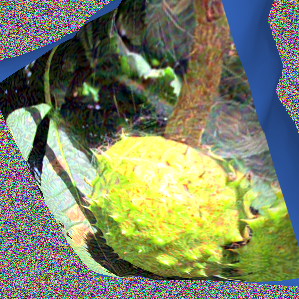}
         \caption{All angles $+25\degree$}
     \end{subfigure}
         \begin{subfigure}[b]{0.155\textwidth}
         \centering
         \includegraphics[width=\textwidth,trim={0cm 0cm 0cm 0cm},clip]{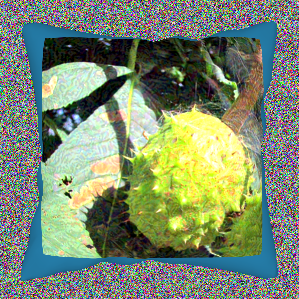}
         \caption{Distance $\times1.2$}
     \end{subfigure}
        }
             \resizebox{0.4\textwidth}{!}{%
\begin{subfigure}[b]{0.155\textwidth}
         \centering
         \includegraphics[width=\textwidth,trim={0cm 0cm 0cm 0cm},clip]{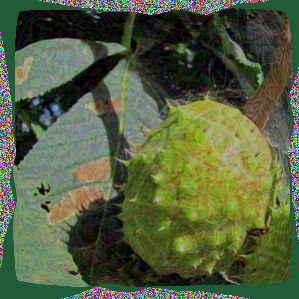}
         \caption{No diffuse light}
     \end{subfigure}
         \begin{subfigure}[b]{0.155\textwidth}
         \centering
         \includegraphics[width=\textwidth,trim={0cm 0cm 0cm 0cm},clip]{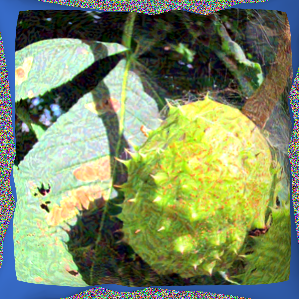}
         \caption{Light on left side}
     \end{subfigure}
         \begin{subfigure}[b]{0.155\textwidth}
         \centering
         \includegraphics[width=\textwidth,trim={0cm 0cm 0cm 0cm},clip]{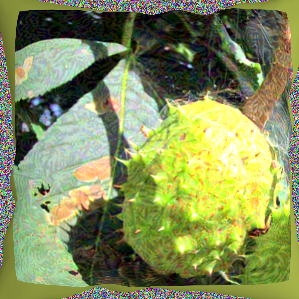}
         \caption{Light on right side}
     \end{subfigure}
        }
        \vspace{-0.2cm}
        \caption{Rendered images with different rendering parameters.}
        \vspace{-0.4cm}
        \label{fig:fig4}
\end{figure}

\section{Methodology}
The object-based diverse input method preprocesses images before feeding them to the source network during the iterative optimization of adversarial examples. Since the ODI method uses a differentiable renderer, even if the rendered image is fed into the network, the gradient of the adversary's objective loss with respect to the input image can be computed via back-propagation. The overall pipeline of the ODI method is illustrated in~\cref{fig:fig3}, and the detailed algorithm of the ODI method is described in the supplementary material. The entire procedure of the ODI method can be broken down into three stages, which we will cover in detail. 

\noindent\textbf{Preparation of an adversarial 3D mesh.} 
The ODI method employs a 3D object as a canvas to draw an adversarial example. Thus, it does not matter if the object changes during iterations. This is where the ODI method significantly differs from previous studies \cite{athalye2018synthesizing,zeng2019adversarial,xiao2019meshadv} that employ 3D meshes for generating physical adversarial examples. First, we randomly select an object from the source object pool. The selected object has a triangular mesh, a texture map, and a bounding box that indicates the canvas region on which the adversarial example will be drawn in the texture map. Next, we fill the texture map with a random solid color, and then we resize and insert the adversarial example into the texture map's bounding box region. Within the frame area, we can also leverage existing input diversification techniques, such as RDI, but we exclude them to clearly demonstrate the effectiveness of ODI in comparison to existing approaches.

\noindent\textbf{Rendering environment setup.} The rendering environment includes lighting and cameras, which are required to render 3D objects. For the camera, we fix the intrinsic parameters and adjust the extrinsic parameters. We alter the three camera angles: elevation, azimuth, and tilt. Please refer to Fig. \ref{fig:fig3} for the definition of each camera angle, and Fig. \ref{fig:fig4} for demonstrations of how each angle alters the viewpoint. In the ODI method, the 3D mesh is initially scaled so that the projected image occupies about $85\%$ of the rendered image in the default view. The three camera angles and camera distance are randomly sampled within a preset range. %The camera moves within the spherical surface, maintaining a distance from the object.

There are two primary types of lights for illumination models ---  directional lights and point lights. We employ point lights in our work, but directional lights can also be used. We randomly adjust the brightness of ambient and diffuse light within a preset range. Additionally, we alter the position of the light by adding a random displacement to its base position, causing it to be randomly placed within a box. \Cref{fig:fig4} illustrates examples of rendered images with different lighting. 

\noindent\textbf{Rendering and blending with backgrounds.} Finally, we render the adversarial 3D mesh in the sampled environment and blend it with a randomly generated background image to create the final output image. This image will be fed into the source network, which \textit{aids in optimizing the input image to appear as the target class in a wide variety of contexts, enhancing transferability}. We present the algorithm of the ODI-MI-TI-FGSM method in \cref{alg:odi}.

\begin{algorithm}[t]
    \algnewcommand\algorithmicinput{\textbf{Input:}}
    \algnewcommand\Input{\item[\algorithmicinput]}
    \algnewcommand\algorithmicoutput{\textbf{Output:}}
    \algnewcommand\Output{\item[\algorithmicoutput]}
    \caption{ODI-MI-TI-FGSM}
    \label{alg:odi}
    \begin{flushleft}
    \textbf{Input:} A clean example $\vct x$; a target label $y_{t}$; a classifier $f$.\\
    \textbf{Input:} Adversary's loss function $\mathcal{L}$; $\ell_\infty$ perturbation constraint $\epsilon$; step size $\alpha$; maximum iterations $T$; decay factor $\mu$; and Gaussian kernel $\mW$.\\
    \textbf{Output:} An adversarial example $\vct x^{adv}$
\end{flushleft}
	\begin{algorithmic}[1]
		\State $\vct g_0 = 0$; $\vct x_0^{adv}=\vct x$
		\For{$t = 0 \rightarrow T-1$}
		    \State Calculate the gradient $\hat{\vct g}_{t+1}$ \Comment{Apply ODI}
		    \begin{equation}
		        \hat{\vct g}_{t+1} = \bigtriangledown_{\vct x^{adv}_t}\mathcal{L}(f(ODI(\vct x^{adv}_t)), y_{t})
		    \end{equation}
		   \State $\tilde{\vct g}_{t+1} = \mu \cdot \vct g_t + \frac{\hat{\vct g}_{t+1}}{\|\hat{\vct g}_{t+1}\|_1}$ \Comment{Apply MI}
		  \State $\vct g_{t+1}=\mW \ast \tilde{\vct g}_{t+1}$ \Comment{Apply TI}
		    \State $\vct x_{t+1}^{adv} = \vct x_t^{adv} - \alpha \cdot \text{sign}(\vct g_{t+1})$ \Comment{Apply FGSM}
		     \State $\vct x_{t+1}^{adv}=Clip_{\vct x}^{\epsilon}(\vct x_{t+1}^{adv})$
		\EndFor
		\State $\vct x^{adv}=\vct x_T^{adv}$
        \State \Return $\vct x^{adv}$
	\end{algorithmic} 
\end{algorithm} 

\renewcommand{\arraystretch}{0.97}

\section{Experiments and Discussion}
\subsection{Experimental Settings}

\begin{table*}[t]
\resizebox{\textwidth}{!}{%
\begin{tabular}{lcccccccccc}
\toprule[0.15em]
\textbf{Source : RN-50} & \multicolumn{9}{c}{Target model} &  \\
\cmidrule(l){2-10}
Attack & VGG-16 & RN-18 & DN-121 & Inc-v3 & Inc-v4 & Mob-v2 & IR-v2 & Adv-Inc-v3 & \begin{tabular}[c]{@{}c@{}}Ens-adv-\\ IR-v2\end{tabular} & \begin{tabular}[c]{@{}c@{}}Computation time\\ per image (sec)\end{tabular} \\ \midrule
DI-MI-TI & 62.2 & 54.9 & 71.4 & 10.5 & 9.0 & 28.5 & 4.5 & 0.0 & 0.0 & 2.7 \\
RDI-MI-TI & 67.8 & 73.4 & 82.9 & 32.4 & 24.6 & 44.5 & 17.4 & 0.0 & 0.0 & 2.3 \\
RDI-MI-TI-SI & 71.2 & 81.7 & 88.5 & 56.6 & 42.8 & 58.0 & 36.6 & 0.2 & 0.9 & 11.2 \\
RDI-MI-TI-VT & 70.3 & 78.7 & 82.5 & 44.6 & 39.1 & 54.4 & 33.7 & 0.2 & 1.7 & 14.1 \\\rowcolor{Gray}
ODI-MI-TI & 76.8 & 77.0 & 86.8 & 67.4 & 55.4 & 66.8 & 48.0 & 0.7 & 1.7 & 6.0 \\\rowcolor{Gray}
ODI-MI-TI-VT & \textbf{81.6} & \textbf{84.4} & \textbf{89.2} & \textbf{74.5} & \textbf{65.9} & \textbf{75.6} & \textbf{62.3} & \textbf{4.6} & \textbf{8.7} & 57.2\\
\midrule[0.1em]
\textbf{Source : VGG-16} & \multicolumn{9}{c}{Target model} &  \\
\cmidrule(l){2-10}
Attack & RN-18 & RN-50 & DN-121 & Inc-v3 & Inc-v4 & Mob-v2 & IR-v2 & Adv-Inc-v3 & \begin{tabular}[c]{@{}c@{}}Ens-adv-\\ IR-v2\end{tabular} & \begin{tabular}[c]{@{}c@{}}Computation time\\ per image (sec)\end{tabular} \\ \midrule
DI-MI-TI & 7.6 & 11.2 & 12.7 & 0.6 & 2.3 & 6.3 & 0.2 & 0.0 & 0.0 & 6.1 \\
RDI-MI-TI & 28.7 & 31.5 & 35.9 & 6.6 & 9.5 & 18.0 & 3.7 & 0.0 & 0.0 & 5.4 \\
RDI-MI-TI-SI & 48.0 & 45.6 & 55.2 & 20.1 & 21.0 & 28.4 & 9.9 & 0.0 & 0.0 & 26.2 \\
RDI-MI-TI-VT & 42.5 & 35.5 & 44.3 & 13.4 & 19.0 & 23.4 & 8.3 & 0.0 & 0.0 & 31.8 \\\rowcolor{Gray}
ODI-MI-TI & 60.8 & 64.3 & 71.1 & 37.0 & 38.0 & 47.0 & 21.1 & 0.0 & 0.0 & 9.0 \\\rowcolor{Gray}
ODI-MI-TI-VT & \textbf{72.3} & \textbf{72.0} & \textbf{76.6} & \textbf{48.7} & \textbf{47.9} & \textbf{57.6} & \textbf{34.7} & \textbf{0.4} & \textbf{0.7} & 71.9\\
\midrule[0.1em]
\textbf{Source : DN-121} & \multicolumn{9}{c}{Target model} &  \\
\cmidrule(l){2-10}
Attack & VGG-16 & RN-18 & RN-50 & Inc-v3 & Inc-v4 & Mob-v2 & IR-v2 & Adv-Inc-v3 & \begin{tabular}[c]{@{}c@{}}Ens-adv-\\ IR-v2\end{tabular} & \begin{tabular}[c]{@{}c@{}}Computation time\\ per image (sec)\end{tabular} \\ \midrule
DI-MI-TI & 38.3 & 30.1 & 43.6 & 7.1 & 7.7 & 13.7 & 4.5 & 0.0 & 0.0 & 2.8 \\
RDI-MI-TI & 41.9 & 45.3 & 55.9 & 21.0 & 19.1 & 21.8 & 12.9 & 0.0 & 0.0 & 2.5 \\
RDI-MI-TI-SI & 44.2 & 54.5 & 59.7 & 34.7 & 24.9 & 26.9 & 22.6 & 0.2 & 0.5 & 12.1 \\
RDI-MI-TI-VT & 49.1 & 56.1 & 63.3 & 32.1 & 28.8 & 29.4 & 25.6 & 0.3 & 0.8 & 15.2 \\\rowcolor{Gray}
ODI-MI-TI & 64.5 & 63.4 & 71.6 & 53.5 & 46.4 & 44.2 & 38.3 & 0.4 & 0.7 & 6.2 \\\rowcolor{Gray}
ODI-MI-TI-VT & \textbf{70.7} & \textbf{74.6} & \textbf{79.1} & \textbf{64.1} & \textbf{57.5} & \textbf{57.8} & \textbf{53.0} & \textbf{2.3} & \textbf{5.0} & 71.7\\
\midrule[0.1em]
\textbf{Source : Inc-v3} & \multicolumn{9}{c}{Target model} &  \\
\cmidrule(l){2-10}
Attack & VGG-16 & RN-18 & RN-50 & DN-121 & Inc-v4 & Mob-v2 & IR-v2 & Adv-Inc-v3 & \begin{tabular}[c]{@{}c@{}}Ens-adv-\\ IR-v2\end{tabular} & \begin{tabular}[c]{@{}c@{}}Computation time\\ per image (sec)\end{tabular} \\ \midrule
DI-MI-TI & 4.2 & 2.2 & 3.6 & 5.4 & 4.3 & 2.4 & 3.6 & 0.0 & 0.0 & 2.2 \\
RDI-MI-TI & 3.2 & 4.4 & 4.4 & 6.9 & 8.3 & 3.0 & 5.5 & 0.0 & 0.0 & 1.9 \\
RDI-MI-TI-SI & 4.2 & 7.2 & 6.3 & 10.3 & 10.5 & 5.0 & 11.4 & 0.2 & 0.3 & 9.2 \\
RDI-MI-TI-VT & 4.8 & 8.5 & 8.9 & 11.9 & 14.6 & 6.2 & 12.8 & 0.2 & 0.0 & 11.8 \\ \rowcolor{Gray}
ODI-MI-TI & 15.7 & 14.7 & 17.4 & 30.4 & 32.1 & 14.1 & 26.9 & 0.3 & 0.6 & 5.5 \\\rowcolor{Gray}
ODI-MI-TI-VT & \textbf{26.7} & \textbf{29.1} & \textbf{34.0} & \textbf{52.5} & \textbf{50.8} & \textbf{25.4} & \textbf{45.8} & \textbf{1.9} & \textbf{3.3} & 62.6\\
            \bottomrule%[0.15em]
\end{tabular}%
}
\vspace{-0.2cm}
\caption{Targeted attack success rates (\%) against nine black-box target models with the four source models. For each attack, we also reported the average computation time to generate an adversarial example.}
\label{tab:table1}
\vspace{-0.3cm}
\end{table*}
\noindent\textbf{Dataset and general settings.}
We utilized the DEV set of the ImageNet-Compatible dataset\footnote{\url{https://github.com/cleverhans-lab/cleverhans/tree/master/cleverhans_v3.1.0/examples/nips17_adversarial_competition/dataset}}, which has been widely used in previous works~\cite{li2020towards,zhao2020success}. This dataset provides 1,000 $299{\times}299$-sized images with their target classes. We adopted the widely used $\ell_\infty$-norm perturbation constraint $\epsilon=16/255$. Following \cite{zhao2020success}, we used the step size $\alpha=2/255$ for the iterative attacks. Our approach and all baselines leveraged the simple logit loss (\cref{eqn:logit}) proposed in \cite{zhao2020success} which is superior in targeted attacks. Following \cite{zhao2020success}, each iterative attack method runs for 300 iterations (\ie, $T=300$). 
Each iterative attack was performed using a single NVIDIA RTX 2080Ti GPU.

\begin{figure}[t]
     \centering
         \includegraphics[width=0.235\textwidth,trim={0.3cm 0.3cm 0.3cm 0.3cm},clip]{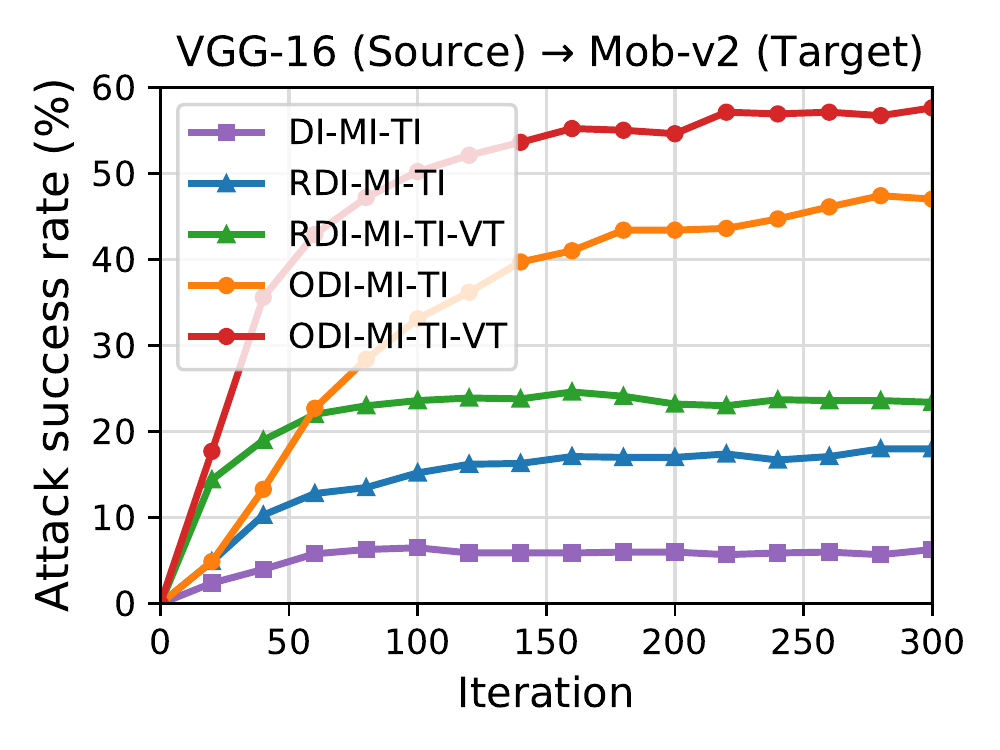}
         \includegraphics[width=0.235\textwidth,trim={0.3cm 0.3cm 0.3cm 0.3cm},clip]{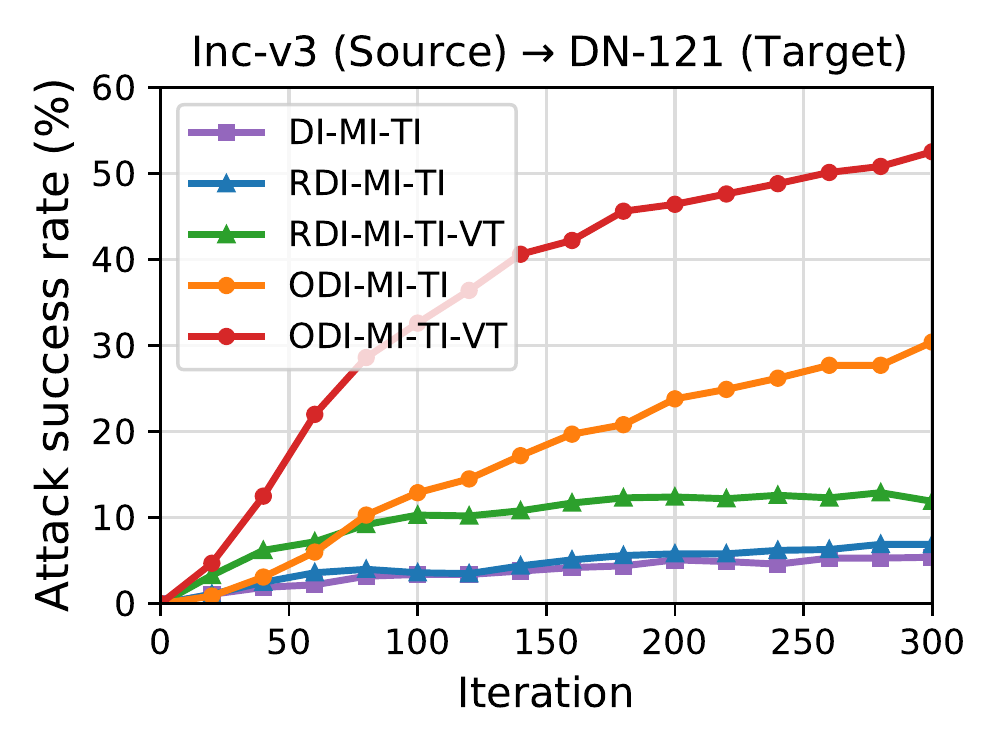}
        \vspace{-0.5cm}
        \caption{Targeted attack success rates (\%) according to the number of iterations.}
        \label{fig:fig5}
        \vspace{-0.4cm}
\end{figure}

\begin{table*}[t]
\resizebox{\textwidth}{!}{%
\begin{tabular}{lccccccccc}
\toprule[0.15em]
\textbf{Source : DN-121} & \multicolumn{9}{c}{Target model} \\\cmidrule(l){2-10}
Source object pool & VGG-16 & RN-18 & RN-50 & Inc-v3 & Inc-v4 & Mob-v2 & IR-v2 & Adv-Inc-v3 & \begin{tabular}[c]{@{}c@{}}Ens-adv-\\ IR-v2\end{tabular} \\\midrule
\{\textit{Package}\} & 59.8 & 54.8 & 65.6 & 43.5 & 37.6 & 35.8 & 29.8 & 0.1 & 0.5 \\
\{\textit{Cup}\} & 34.2 & 45.7 & 47.9 & 37.8 & 29.8 & 28.5 & 26.7 & \textbf{0.5} & \textbf{1.2} \\
\{\textit{Pillow}\} & 64.1 & 57.6 & 68.6 & 44.9 & 40.6 & 39.4 & 30.5 & 0.1 & 0.6 \\
\{\textit{T-shirt}\} & 23.5 & 36.4 & 38.1 & 31.2 & 23.5 & 19.3 & 19.4 & \textbf{0.5} & 1.0 \\
\{\textit{Ball}\} & 46.1 & 26.3 & 36.7 & 17.1 & 16.6 & 17.7 & 10.6 & 0.0 & 0.0 \\
\{\textit{Book}\} & 50.9 & 61.9 & 67.0 & 52.8 & 39.5 & 39.5 & 36.6 & 0.3 & 1.0 \\
\{All 6 objects\} & 60.2 & 59.5 & 66.3 & 48.8 & 42.7 & 42.4 & 35.7 & 0.4 & 0.9 \\ \rowcolor{Gray}
\{\textit{Package, Pillow, Book}\} & \textbf{64.5} & \textbf{63.4}& \textbf{71.6} & \textbf{53.5} & \textbf{46.4} & \textbf{44.2} & \textbf{38.3} & 0.4 & 0.7\\
\bottomrule%[0.12em]
\end{tabular}%
}
\vspace{-0.2cm}
\caption{Targeted attack success rates (\%) of ODI-MI-TI against nine black-box target models with different source object pools. The ensembles of multiple source objects outperform their single object counterparts.
}
\label{tab:table2}
\vspace{-0.3cm}
\end{table*}
\noindent\textbf{Source and target models.}
For fair comparison with existing works, we adopted four models used in \cite{zhao2020success} as source and target models in our experiments --- ResNet-50 (RN-50) \cite{he2016deep}, Inception-v3 (Inc-v3)  \cite{szegedy2016rethinking}, DenseNet-121 (DN-121) \cite{huang2017densely}, and VGG-16\_bn (VGG-16)  \cite{simonyan2014very}. Additionally, we added six additional models in the collection of target models for more comprehensive comparisons --- ResNet-18 (RN-18) \cite{he2016deep}, Inception-v4 (Inc-v4) \cite{szegedy2017inception}, MobileNet-v2 (Mob-v2) \cite{sandler2018mobilenetv2}, Inception ResNet-v2 (IR-v2) \cite{szegedy2017inception}, adversarially trained Inc-v3 (Adv-Inc-v3) \cite{DBLP:journals/corr/abs-1804-00097}, and ensemble-adversarially trained IR-v2 (Ens-adv-IR-v2) \cite{DBLP:journals/corr/abs-1804-00097}.
This paper focuses on single-source model-based transfer attacks to demonstrate the ODI method's effectiveness in a challenging environment. However, we believe that the ensemble of source models can further boost the transfer success rates.

\noindent\textbf{Baselines.}
We employed four baseline attack methods, which are various combinations of six existing techniques: DI \cite{xie2019improving}, RDI \cite{zou2020improving}, MI \cite{dong2018boosting}, TI \cite{dong2019evading}, SI \cite{lin2019nesterov}, and VT \cite{wang2021enhancing}. Please note that the previously reported state-of-the-art method is DI-MI-TI with simple logit loss \cite{zhao2020success}. However, we further improved this method by replacing DI with RDI, combining it with the recently proposed VT and SI, and using them as the state-of-the-art baselines (RDI-MI-TI-SI and RDI-MI-TI-VT).
The maximally enlarged image sizes of DI and RDI were set to $330\times330$ and $340\times340$, respectively. Following \cite{zhao2020success}, the convolution kernel size for TI was set to 5 and $p$ for DI to $0.7$, while the decay factor $\mu$ for MI was set to 1.0. The numbers of scales and samples for SI and VT were set to 5, and $\beta$ for VT to $1.5$. 

We also implemented ATTA-MI-TI, but ATTA \cite{wu2021improving} was initially designed for non-targeted attacks with few iterations, so it performed poorly. Although we changed ATTA to use targeted adversarial examples using logit loss for training, its results were not comparable. We hypothesize that the repeated use of the fixed transformation limits transferability. For a rich comparison, we included the results of MI-TI and ATTA-MI-TI in supplementary material.

\begin{figure}[t]
     \centering
     \begin{subfigure}[b]{0.14\textwidth}
         \centering
         \includegraphics[width=\textwidth,trim={0cm 0cm 0cm 0cm},clip]{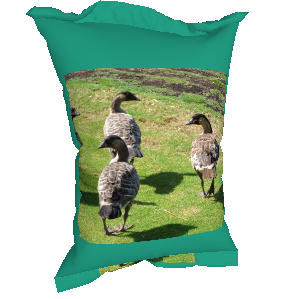}
         \caption{\textit{Package}}
     \end{subfigure}
          \begin{subfigure}[b]{0.14\textwidth}
         \centering
         \includegraphics[width=\textwidth,trim={0cm 0cm 0cm 0cm},clip]{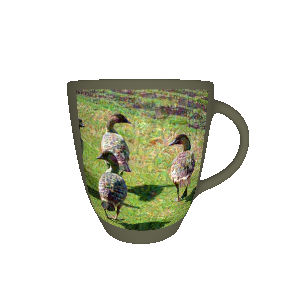}
         \caption{\textit{Cup}}
     \end{subfigure}
          \begin{subfigure}[b]{0.14\textwidth}
         \centering
         \includegraphics[width=\textwidth,trim={0cm 0cm 0cm 0cm},clip]{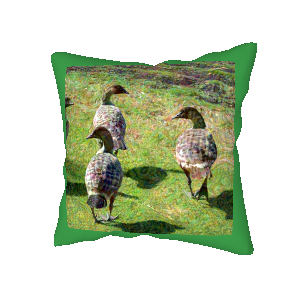}
         \caption{\textit{Pillow}}
     \end{subfigure}
     
          \begin{subfigure}[b]{0.14\textwidth}
         \centering
         \includegraphics[width=\textwidth,trim={0cm 0cm 0cm 0cm},clip]{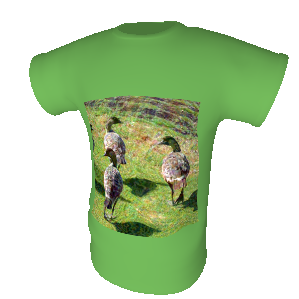}
         \caption{\textit{T-shirt}}
     \end{subfigure}
          \begin{subfigure}[b]{0.14\textwidth}
         \centering
         \includegraphics[width=\textwidth,trim={0cm 0cm 0cm 0cm},clip]{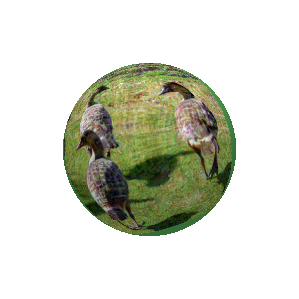}
         \caption{\textit{Ball}}
     \end{subfigure}
          \begin{subfigure}[b]{0.14\textwidth}
         \centering
         \includegraphics[width=\textwidth,trim={0cm 0cm 0cm 0cm},clip]{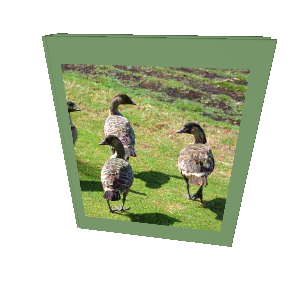}
         \caption{\textit{Book}}
     \end{subfigure}
        \caption{Six source objects used in our experiments. An image is printed on them to visualize the area of adversarial texture.}
        \label{fig:fig6}
        \vspace{-0.4cm}
\end{figure}

\noindent \textbf{Settings for the ODI method.}
We utilized the PyTorch3D library \cite{ravi2020accelerating} for the differentiable rendering in the ODI method. For the parameters of the ODI method, we constructed the source object pool as \{\textit{Package, Pillow, Book}\}, each of which is shown in~\cref{fig:fig6}. The ranges of the solid texture color, camera angles and distance were set to $[0.1,0.7]$, $[-35\degree, 35\degree]$, and $[0.8, 1.2]$, respectively. The default position of the light was set to $[0, 0, 4]$, and its maximum displacement to $2$. 
The brightness range of ambient light was set to $[0.6,0.9]$ and the brightness range of diffuse light to $[0, 0.5]$. Finally, we set the shininess connected to the material's reflectance to $0.5$.

\begin{table*}[t]
\centering

\resizebox{\textwidth}{!}{%
\begin{tabular}{ccccccccccc}
\toprule[0.15em]
\multicolumn{1}{l}{\textbf{Source : Inc-v3}} &  & \multicolumn{9}{c}{Target model} \\ \cmidrule(l){3-11}
Ablation & \multicolumn{1}{c}{Value} & \multicolumn{1}{c}{VGG-16} & \multicolumn{1}{c}{RN-18} & \multicolumn{1}{c}{RN-50} & \multicolumn{1}{c}{DN-121} & \multicolumn{1}{c}{Inc-v4} & \multicolumn{1}{c}{Mob-v2} & \multicolumn{1}{c}{IR-v2} & \multicolumn{1}{c}{Adv-Inc-v3} & \multicolumn{1}{c}{\begin{tabular}[c]{@{}c@{}}Ens-adv-\\ IR-v2\end{tabular}} \\\midrule
\multirow{5}{*}{Angle} & $-5\degree\sim 5\degree$ & 5.4 & 4.1 & 4.8 & 8.1 & 7.8 & 3.7 & 7.0 & 0.2 & 0.0 \\
 &$-15\degree\sim 15\degree$ & 9.1 & 7.7 & 10.9 & 19.4 & 20.7 & 7.4 & 16.7 & 0.0 & 0.1 \\
 & $-25\degree\sim 25\degree$  & 13.4 & 11.9 & 14.2 & 28.7 & 28.6 & 11.3 & 22.7 & 0.2 & 0.0 \\
 &\cellcolor{Gray}$-35\degree\sim 35\degree$  & \cellcolor{Gray}\textbf{15.6} & \cellcolor{Gray}\textbf{14.1} & \cellcolor{Gray}\textbf{17.9} & \cellcolor{Gray}\textbf{31.6} & \cellcolor{Gray}\textbf{30.5} &\cellcolor{Gray} 12.9 &\cellcolor{Gray}\textbf{24.9} & \cellcolor{Gray}\textbf{0.3} &\cellcolor{Gray}\textbf{0.4} \\
 & $-45\degree\sim 45\degree$  & 15.4 & \textbf{14.1} & 16.9 & 30.1 & 26.0 & \textbf{13.2} & 22.2 & 0.2 & 0.1 \\
 \midrule
\multirow{3}{*}{Distance} & \cellcolor{Gray}$0.8\times \sim 1.2\times$ & \cellcolor{Gray}\textbf{15.6} & \cellcolor{Gray}\textbf{14.1} & \cellcolor{Gray}\textbf{17.9} & \cellcolor{Gray}\textbf{31.6} & \cellcolor{Gray}\textbf{30.5} & \cellcolor{Gray}\textbf{12.9} & \cellcolor{Gray}\textbf{24.9} & \cellcolor{Gray}\textbf{0.3} & \cellcolor{Gray}\textbf{0.4} \\
 &  $0.9\times \sim 1.1\times$ & 12.5 & 12.4 & 14.6 & 26.9 & 26.0 & 11.2 & 21.2 & 0.1 & 0.2 \\
  &  $1.0\times$ & 11.1 & 11.4 & 12.6 & 23.4 &  23.4 & 9.1 & 18.3 & 0.0 & 0.1  \\ \midrule
\multirow{4}{*}{Background} &  \cellcolor{Gray}Random pixel & \cellcolor{Gray}15.6 & \cellcolor{Gray}\textbf{14.1} & \cellcolor{Gray}\textbf{17.9} & \cellcolor{Gray}\textbf{31.6} & \cellcolor{Gray}\textbf{30.5} & \cellcolor{Gray}12.9 & \cellcolor{Gray}\textbf{24.9} & \cellcolor{Gray}\textbf{0.3} & \cellcolor{Gray}\textbf{0.4} \\
 & Random solid & 15.7 & 13.4 & \textbf{16.9} & 30.7 & 27.9 & 12.5 & 22.3 & 0.2 & 0.3 \\
 & Blurred image & \textbf{15.8} & 12.5 & 17.0 & \textbf{31.6} & 28.5 & 12.3 & 22.1 & \textbf{0.3} & 0.3 \\
 & Black solid & 14.7 & 13.1 & 16.2 & 28.1 & 28.4 & \textbf{13.1} & 21.9 & \textbf{0.3} & 0.1\\\bottomrule
\end{tabular}%
}
\vspace{-0.2cm}
\caption{Targeted attack success rates (\%) against nine black-box target models with different camera parameters and backgrounds. For these experiments, we used \textit{Pillow} as the source object.}
\label{tab:table4}
\vspace{-0.4cm}
\end{table*}

\subsection{Experimental Results}
\noindent\textbf{Transfer success rates.} Table \ref{tab:table1} shows the targeted attack success rates against nine black-box target models with the four source models. 
The ODI-MI-TI-VT enhanced the average attack success rate from 28.3\% to 47.0\% compared to the best performance of baselines. The ODI method's performance improvements were most prominent when the source model was VGG-16. 
Compared to DI-MI-TI, which is the previous state-of-the-art technique, ODI-MI-TI and ODI-MI-TI-VT boosted the average transfer success rates from $4.5\%$ to $37.7\%$ $(8\times)$ and $45.7\%$ $(10\times)$, respectively. 
The targeted attack success rates for two cases are shown in~\cref{fig:fig5}, and the rest of the plots can be found in supplementary material.

\noindent\textbf{Computation time.} In Table \ref{tab:table1}, we also reported the average time required to generate an adversarial example. Due to the rendering overhead, the ODI method requires more computational cost than DI and RDI. However, when compared to RDI-MI-TI-VT and RDI-MI-TI-SI,
ODI-MI-TI cut the required time by half and increased the attack success rate by $10.8\%$ on average. When VT or SI is used, the amount of computing required increases significantly because each loop requires more inference in proportion to the numbers of scales and samples.

\noindent\textbf{Attacks on adversarially trained models.} 
Adversarial training is widely considered to be one of the most effective adversarial defenses. Targeted attacks on adversarially trained models \cite{madry2018towards,tramer2018ensemble} are more challenging in the black-box setting. Hence, the previously reported state-of-the-art method, DI-MI-TI, recorded $0\%$ attack success rate on Adv-Inc-v3 \cite{madry2018towards, DBLP:journals/corr/abs-1804-00097} and Ens-adv-IR-v2 \cite{tramer2018ensemble,DBLP:journals/corr/abs-1804-00097}. Surprisingly, ODI-MI-TI-VT showed up to $4.6\%$ and $8.7\%$ attack success rates against these models. Given the significant architectural differences between the source and target models, this improvement is noteworthy.

\subsection{Ablation Studies}
In this section, we conduct extensive ablation studies on the proposed approach and describe our findings.

\noindent\textbf{Variation of source object pools.}
Since the 3D source object in the ODI method is like a canvas for adversarial examples, the object can be freely selected from various 3D objects. As candidate source models, we chose six objects\footnote{We included references to these 3D models in supplementary material.} that are commonly observed in our daily life, as shown in~\cref{fig:fig6}. Based on these models, we performed single-object-based and multi-object-based attacks. The results are summarized in Table \ref{tab:table2}. Interestingly, among single-object-based attacks, \textit{Pillow} had the highest attack success rate against VGG-16, whereas \textit{Book} recorded the highest success rate against Inc-v3. This means that an object's transferability enhancement varies depending on the target model. Moreover, the results of multi-object-based attacks 
reveal that the transferability enhancements of each 3D object complement each other. Finally, an ensemble of three objects outperformed the ensemble of all objects, highlighting the importance of carefully assembling the source object pool.

\begin{figure}[t]
     \centering
     \begin{subfigure}[b]{0.11\textwidth}
         \centering
         \includegraphics[width=\textwidth,trim={0cm 0cm 0cm 0cm},clip]{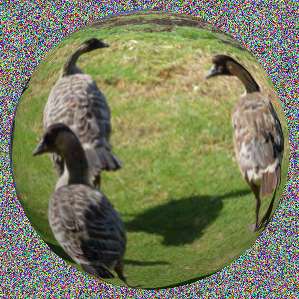}
         \caption{\textit{One ball}}
     \end{subfigure}
          \begin{subfigure}[b]{0.11\textwidth}
         \centering
         \includegraphics[width=\textwidth,trim={0cm 0cm 0cm 0cm},clip]{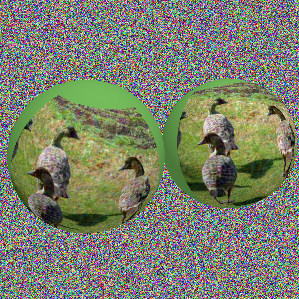}
         \caption{\textit{Two balls}}
     \end{subfigure}
          \begin{subfigure}[b]{0.11\textwidth}
          %\vspace{0.1cm}
         \centering
         \includegraphics[width=\textwidth,trim={0cm 0cm 0cm 0cm},clip]{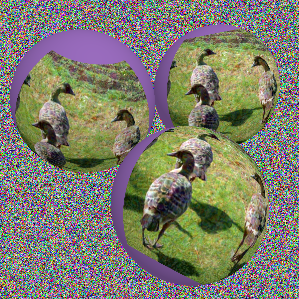}
         \caption{\textit{Three balls}}
     \end{subfigure}
          \begin{subfigure}[b]{0.11\textwidth}
         \centering
         \includegraphics[width=\textwidth,trim={0cm 0cm 0cm 0cm},clip]{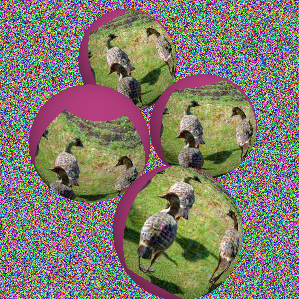}
         \caption{\textit{Four balls}}
     \end{subfigure}
        \vspace{-0.2cm}
        \caption{Rendered images of different numbers of balls.}
        \label{fig:fig7}
        \vspace{-0.2cm}
\end{figure}

\noindent\textbf{Variation of the number of objects.}
Instead of rendering a single object, we can visualize multiple  objects to diversify the input. Using the balls as shown in Fig. \ref{fig:fig7}, we study the transferability of the adversarial example as a function of the number of objects. The attack success rate for each target model can vary based on the number of source objects, even though the same 3D object is used. When the source model was VGG-16, \textit{Three balls} showed the highest success rates for most cases, but adversarially trained models were most vulnerable to \textit{Four balls} rather than \textit{Three balls}. Detailed results are included in supplementary material.

\begin{figure}[t]
     \centering
     \resizebox{0.35\textwidth}{!}{%
     \begin{subfigure}[b]{0.13\textwidth}
         \centering
         \includegraphics[width=\textwidth,trim={0cm 0cm 0cm 0cm},clip]{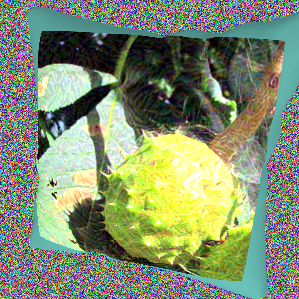}
         \caption{Random pixel}
     \end{subfigure}
          \begin{subfigure}[b]{0.13\textwidth}
         \centering
         \includegraphics[width=\textwidth,trim={0cm 0cm 0cm 0cm},clip]{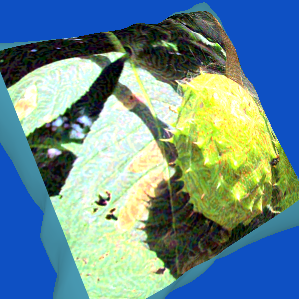}
         \caption{Random solid}
     \end{subfigure}
          \begin{subfigure}[b]{0.13\textwidth}
         \centering
         \includegraphics[width=\textwidth,trim={0cm 0cm 0cm 0cm},clip]{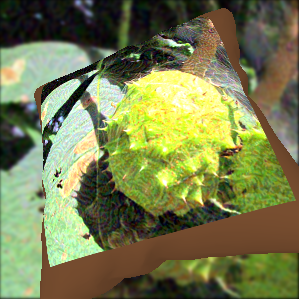}
         \caption{Blurred image}
     \end{subfigure}
     }
        \vspace{-0.2cm}
        \caption{Illustrations of different backgrounds.}
        
        \label{fig:fig8}
        \vspace{-0.5cm}
\end{figure}
\noindent\textbf{Variation of the camera angle and distance.}
We conducted experiments to assess the impact of camera angle and distance variations. The attack success rate increased proportionally with the camera angle variation but dropped beyond a certain range. This finding suggests that excessive image transformation may deteriorate the transferability as the adversarial examples overfit to the source model's drastic transformation. Table \ref{tab:table4} shows that the attack success rate increases as the range of the camera distance expands.
Therefore, we expect that the more rendering parameters vary, the better the transferability, within a certain range.

\noindent\textbf{Different backgrounds.} Finally, we examined the effect of the backgrounds. In this experiment, we employed a random solid background, random pixel values, and a blurred image. All random values were uniformly sampled between 0 and 1. For blurred images, we convolved the input image with the Gaussian kernel whose kernel size is $50$ and $\sigma$ is $15$. According to the experimental results, the above three backgrounds performed better than the solid black background. Among all, the background with random pixel values obtained the highest attack success rate.

\begin{figure}[t]
     \centering
     \begin{subfigure}[b]{0.37\textwidth}
         \centering
         %\textbf{Dodging Attack}\\\smallskip
         \includegraphics[width=\textwidth,trim={0cm 0.4cm 0cm 0cm},clip]{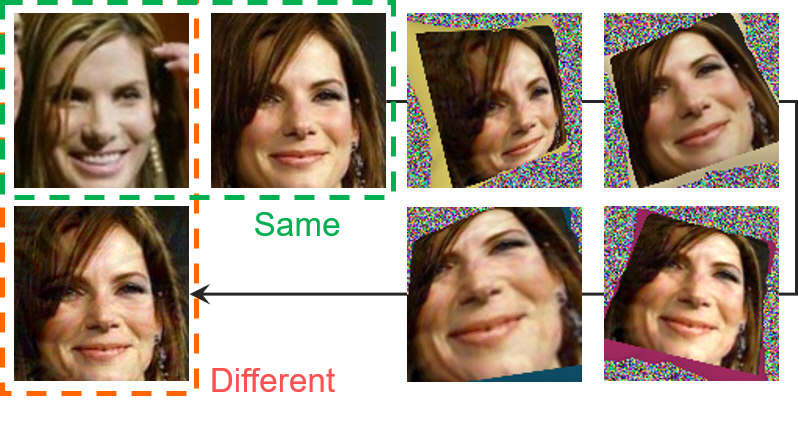}
         \caption{Dodging attack}
     \end{subfigure}
     
          \begin{subfigure}[b]{0.37\textwidth}
         \centering
         %\textbf{Impersonation Attack}\\\smallskip
         \includegraphics[width=\textwidth,trim={0cm 0.4cm 0cm 0cm},clip]{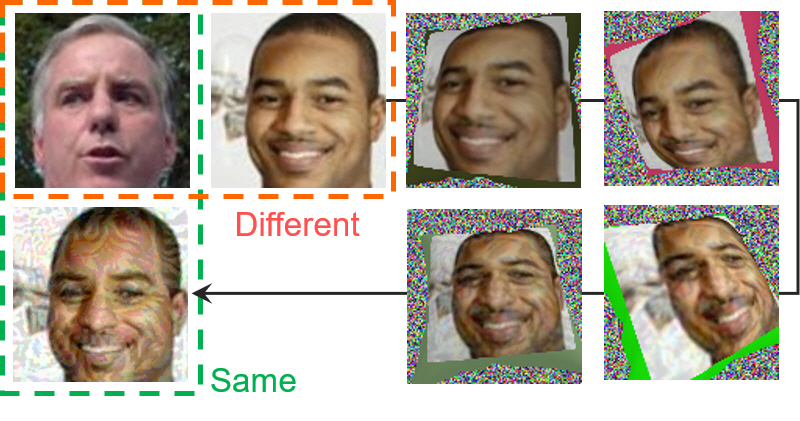} 
         \caption{Impersonation attack}
     \end{subfigure}
        \vspace{-0.2cm}
        \caption{Illustrations of adversarial attacks with the ODI method on the face verification task.}
        \label{fig:fig9}
        \vspace{-0.4cm}
\end{figure}

\subsection{Adversarial Attacks on Face Recognition}
The proposed technique is not limited to image classification. It can be used for adversarial attacks on other tasks as well, including face recognition \cite{deng2019arcface} and object detection \cite{redmon2016you}.
As an example, we applied our method to adversarial attacks on face verification models. 
A face verification model compares two photos to see if they are of the same person \cite{deng2019arcface}. By changing an image in the pair of photos, adversaries can launch two types of attacks, which cause the same person to be classified as different people (dodging attacks) or different people to be classified as the same person (impersonation attacks) \cite{dong2019efficient}. These two types of attacks with the ODI method are illustrated in~\cref{fig:fig9}.%%Further information on these attacks can be found in related studies.

We experimented with 500 face pairs in the Labeled Faces in the Wild (LFW) dataset \cite{LFWTech} which is widely used in related works \cite{schroff2015facenet,deng2019arcface,huang2020curricularface}. We used the squared $\ell_2$-distance between a pair of facial features as the adversary's objective loss, $\epsilon=8/255$ and $\epsilon=16/255$ for dodging and impersonation attacks, respectively. We used \textit{Pillow} as the source object and changed the range of camera angles to $[-25\degree,25\degree]$. We used the same setting of the ImageNet dataset for all other parameters to show that a significant performance gain is attainable without expensive parameter searches. The attack success rates on three black-box  models from the ArcFace ResNet-50 source model \cite{deng2019arcface} are shown in Table \ref{tab:table5}. ODI-MI-TI-VT boosted the attack success rate for impersonation and dodging attacks by an average of $12.0\%$ and $7.2\%$, respectively. 

\begin{table}[t]

\resizebox{0.47\textwidth}{!}{%
\begin{tabular}{lccc}
\toprule[0.15em]
 Impersonation attack & \multicolumn{3}{c}{Target model} \\ \cmidrule(l){2-4}  
Method & \begin{tabular}[c]{@{}c@{}}CurricularFace \cite{huang2020curricularface}\\ RN-100  \cite{he2016deep}\end{tabular} & \begin{tabular}[c]{@{}c@{}}ArcFace \cite{deng2019arcface}\\ GhostNet x1.3~\cite{han2020ghostnet}\end{tabular} & FaceNet \cite{schroff2015facenet} \\
    \midrule
\multicolumn{1}{l}{DI-MI-TI} & 57.0 & 47.2 & 65.2 \\
\multicolumn{1}{l}{RDI-MI-TI} & 58.4 & 47.8 & 68.0 \\
\multicolumn{1}{l}{RDI-MI-TI-VT} & 61.4 & 51.4 & 71.4 \\ \rowcolor{Gray}
\multicolumn{1}{l}{ODI-MI-TI} & 71.8 & 61.8 & 80.6 \\ \rowcolor{Gray}
\multicolumn{1}{l}{ODI-MI-TI-VT} & \textbf{73.0} & \textbf{63.0} & \textbf{84.2} \\ 
\midrule[0.1em]
 Dodging attack& \multicolumn{3}{c}{Target model} \\ \cmidrule(l){2-4} 
Method & \begin{tabular}[c]{@{}c@{}}CurricularFace \cite{huang2020curricularface}\\ RN-100 \cite{he2016deep}\end{tabular} & \begin{tabular}[c]{@{}c@{}}ArcFace\cite{deng2019arcface}\\ GhostNet x1.3~\cite{han2020ghostnet}\end{tabular} & FaceNet \cite{schroff2015facenet}\\
    \midrule
\multicolumn{1}{l}{DI-MI-TI} & 97.2 & 62.4 & 90.8 \\
\multicolumn{1}{l}{RDI-MI-TI} & 96.2 & 68.4 & 90.6 \\
\multicolumn{1}{l}{RDI-MI-TI-VT} & 97.0 & 71.2 & 92.0 \\ \rowcolor{Gray}
\multicolumn{1}{l}{ODI-MI-TI} & 99.2 & 83.0 & 95.2 \\ \rowcolor{Gray}
\multicolumn{1}{l}{ODI-MI-TI-VT} & \textbf{99.4} & \textbf{85.2} & \textbf{97.2}\\
    \bottomrule
\end{tabular}%
}
\vspace{-0.2cm}
\caption{Transfer attack success rates (\%) on the face verification models. The source model is ArcFace ResNet-50 \cite{deng2019arcface}.}
\label{tab:table5}
\vspace{-0.4cm}
\end{table}
\subsection{Discussion}
\noindent\textbf{Limitations.}
Our technique is suitable for sufficiently large images that are able to preserve the original image contents with 3D models. It is hard to apply the ODI method to small images, such as the $32{\times}32$-sized images in the CIFAR-10 dataset \cite{krizhevsky2009learning}, unless we expand the input image.

\noindent\textbf{Potential societal impact.}
If the proposed ODI method is maliciously utilized, it may cause social confusion by deceiving AI-based commercial services in the world. However, our study raises awareness of this potential threat to service providers and  researchers, and can contribute to developing more robust deep learning models.

\noindent\textbf{Future work.}
Although this work focuses on rigid objects whose shapes do not change, more diverse augmentation is achievable by altering the objects' shapes. More work has to be done with this concept to increase the transferability even further. Additionally, this 3D model-based data augmentation can be used to improve the performance of deep learning models on other tasks. This technique could be especially useful when the amount of training data is limited.
\section{Conclusion}
In this paper, we have proposed the object-based diverse input method as a novel data augmentation technique to improve the transferability of targeted adversarial examples. Our experimental results on image classification and face verification demonstrate that the proposed method boosts the attack success rate on various black-box target models. % %
%%%%%%%%% REFERENCES
{\small
\bibliographystyle{ieee_fullname}
\bibliography{egbib}
}

\end{document}

\definecolor{Gray}{gray}{0.92}
\title{Supplementary Material:\\ Improving the Transferability of Targeted Adversarial Examples through Object-Based Diverse Input}

\maketitle

%%%%%%%%% ABSTRACT
%\appendix
%\section*{Appendices}
\section*{A. References to used 3D models}
In our experiments, we used six 3D models as source objects. All of them are free downloadable and have royalty-free licenses.
\begin{itemize}
    \item \textit{T-shirt}: \url{https://www.turbosquid.com/3d-models/3ds-max-tshirt-soccer/570329}
    \item \textit{Book}: \url{https://www.turbosquid.com/3d-models/free-3ds-mode-book-modelled/656471}
    \item \textit{Ball}: \url{https://www.turbosquid.com/3d-models/pool-balls-max-free/877865}
    \item \textit{Pillow}: \url{https://www.turbosquid.com/3d-models/3d-decorative-kawaii-pillows-1543064}
    \item \textit{Cup}: \url{https://www.turbosquid.com/3d-models/free-cups-spongebob-3d-model/935414}
    \item \textit{Package}: \url{https://www.cgtrader.com/free-3d-models/food/miscellaneous/lays-packaging-chips}
\end{itemize}
\begin{algorithm*}[t]
   \caption{Object-based Diverse Input (ODI) method.}
	\label{alg:odi}
	\begin{algorithmic}[1]
		\Require An image $\vct x$.
		\Require A source object pool $\set O$; the range of texture color $(l_{texture},u_{texture})$; the range of camera angles $(l_{angle},u_{angle})$; the range of camera distance $(l_{dist},u_{dist})$; the range of ambient light brightness $(l_{ambient},u_{ambient})$;
		the range of diffuse light brightness $(l_{diffuse},u_{diffuse})$; the range of translation of the light $(l_{light},u_{light})$, and the default position of the light $\vct l_d$.
		
		\Ensure The transformed image $\vct x^{\prime\prime}$.
		\State $\vct o \sim \set O$ // Randomly choose an object in the object pool
		\State ($\vct t, \vct b, \vct m) =\vct o$ // Get the texture map $\vct t$, bounding box of adversarial texture $\vct b$, and the mesh $\vct m$ from the object
		\State  $\vct t^{\prime}=Fill(\vct t, \vct c)$, where  $\vct c \sim \set U (l_{texture},u_{texture})$ // Paint the texture $\vct t$ with the random solid color % l^_{texture},u^_{texture})
		\State $\vct t^{\prime}[\vct b] = Resize(\vct x, \vct b)$ // Insert the resized input image $\vct x$ into the area of $\vct b$ of the painted texture $\vct t^{\prime}$
		\State $\vct c_a \sim \set U(l_{angle},u_{angle})$ // Randomly choose the camera angles.
		\State $c_d \sim \set U(l_{dist}, u_{dist})$ // Randomly choose the camera distance.
		\State $\vct l_{a} = l_a\vct 1$, where $l_a  \sim \set U(l_{ambient}, u_{ambient}) $ // Randomly choose the ambient light color.
		\State $\vct l_{d}= l_d\vct 1$, where $l_d \sim \set U(l_{diffuse}, u_{diffuse}) $ // Randomly choose the diffuse light color.
		\State $\vct l_{p} = \vct l_{d} + \vct l_r$ where $\vct l_r \sim \set U(l_{light},u_{light})$  // Randomly choose the translation amount of the light.
		\State $\vct x^{\prime} =Render(\vct t^{\prime}, \vct m, \vct c_a, c_d, \vct l_a, \vct l_d, \vct l_p)$ // Render the modified textured 3D mesh with the randomly chosen parameters.
		\State $\vct b \sim \set U(0, 1)$ // Randomly generate the background image 
        \State $\vct  x^{\prime\prime} = Blend(\vct x^{\prime}, \vct b$)
		\State \Return $\vct x^{\prime\prime}$
	\end{algorithmic}
\end{algorithm*}

%\section{Algorithm}

\begin{table*}[t]
\resizebox{\textwidth}{!}{%
\begin{tabular}{lcccccccccc}
\toprule[0.15em]
\textbf{Source : RN-50} & \multicolumn{9}{c}{Target model} &  \\
\cmidrule(l){2-10}
Attack & VGG-16 & RN-18 & DN-121 & Inc-v3 & Inc-v4 & Mob-v2 & IR-v2 & Adv-Inc-v3 & \begin{tabular}[c]{@{}c@{}}Ens-adv-\\ IR-v2\end{tabular} & \begin{tabular}[c]{@{}c@{}}Computation time\\ per image (sec)\end{tabular} \\ \midrule
MI-TI & 3.2 & 6.6 & 8.4 & 0.3 & 0.4 & 2.4 & 0.1 & 0.0 & 0.0 & 2.3\\
ATTA-MI-TI & 6.6 & 10.8 & 14.7 & 0.9 & 0.8 & 3.8 & 0.2 & 0.0 & 0.0 & 14.1\\
DI-MI-TI & 62.2 & 54.9 & 71.4 & 10.5 & 9.0 & 28.5 & 4.5 & 0.0 & 0.0 & 2.7 \\
RDI-MI-TI & 67.8 & 73.4 & 82.9 & 32.4 & 24.6 & 44.5 & 17.4 & 0.0 & 0.0 & 2.3 \\\rowcolor{Gray}
ODI-MI-TI & \textbf{76.8} & \textbf{77.0} & \textbf{86.8} & \textbf{67.4} & \textbf{55.4} & \textbf{66.8} & \textbf{48.0} & \textbf{0.7} & \textbf{1.7} & 6.0 \\
\midrule[0.1em]
\textbf{Source : VGG-16} & \multicolumn{9}{c}{Target model} &  \\
\cmidrule(l){2-10}
Attack & RN-18 & RN-50 & DN-121 & Inc-v3 & Inc-v4 & Mob-v2 & IR-v2 & Adv-Inc-v3 & \begin{tabular}[c]{@{}c@{}}Ens-adv-\\ IR-v2\end{tabular} & \begin{tabular}[c]{@{}c@{}}Computation time\\ per image (sec)\end{tabular} \\ \midrule
MI-TI & 0.8 & 1.0 & 1.3 & 0.0 & 0.2 & 0.9 & 0.0 & 0.0 & 0.0 & 5.3\\
ATTA-MI-TI & 1.6 & 2.6 & 3.5 & 0.2 & 0.2 & 1.7 & 0.0 & 0.0 & 0.0 & 21.0\\
DI-MI-TI & 7.6 & 11.2 & 12.7 & 0.6 & 2.3 & 6.3 & 0.2 & 0.0 & 0.0 & 6.1 \\
RDI-MI-TI & 28.7 & 31.5 & 35.9 & 6.6 & 9.5 & 18.0 & 3.7 & 0.0 & 0.0 & 5.4 \\ \rowcolor{Gray}
ODI-MI-TI & \textbf{60.8} & \textbf{64.3} & \textbf{71.1} & \textbf{37.0} & \textbf{38.0} & \textbf{47.0} & \textbf{21.1} & 0.0 & 0.0 & 9.0 \\
\midrule[0.1em]
\textbf{Source : DN-121} & \multicolumn{9}{c}{Target model} &  \\
\cmidrule(l){2-10}
Attack & VGG-16 & RN-18 & RN-50 & Inc-v3 & Inc-v4 & Mob-v2 & IR-v2 & Adv-Inc-v3 & \begin{tabular}[c]{@{}c@{}}Ens-adv-\\ IR-v2\end{tabular} & \begin{tabular}[c]{@{}c@{}}Computation time\\ per image (sec)\end{tabular} \\ \midrule
MI-TI & 4.1 & 5.5 & 6.5 & 0.3 & 0.6 & 2.2 & 0.5 & 0.0 & 0.0 & 2.5\\
ATTA-MI-TI & 6.9 & 8.9 & 10.6 & 1.4 & 1.6 & 3.6 & 0.8 & 0.0 & 0.0 & 18.7\\
DI-MI-TI & 38.3 & 30.1 & 43.6 & 7.1 & 7.7 & 13.7 & 4.5 & 0.0 & 0.0 & 2.8 \\
RDI-MI-TI & 41.9 & 45.3 & 55.9 & 21.0 & 19.1 & 21.8 & 12.9 & 0.0 & 0.0 & 2.5 \\\rowcolor{Gray}
ODI-MI-TI & \textbf{64.5} & \textbf{63.4} & \textbf{71.6} & \textbf{53.5} & \textbf{46.4} & \textbf{44.2} & \textbf{38.3} & \textbf{0.4}& \textbf{0.7}& 6.2 \\
\midrule[0.1em]
\textbf{Source : Inc-v3} & \multicolumn{9}{c}{Target model} &  \\
\cmidrule(l){2-10}
Attack & VGG-16 & RN-18 & RN-50 & DN-121 & Inc-v4 & Mob-v2 & IR-v2 & Adv-Inc-v3 & \begin{tabular}[c]{@{}c@{}}Ens-adv-\\ IR-v2\end{tabular} & \begin{tabular}[c]{@{}c@{}}Computation time\\ per image (sec)\end{tabular} \\ \midrule
MI-TI &  0.3 & 0.0 & 0.1 & 0.1 & 0.1 & 0.1 & 0.1 & 0.1 & 0.0 & 1.9\\
ATTA-MI-TI & 0.3 & 0.3 & 0.2 & 0.4 & 0.5 & 0.4 & 0.1 & 0.0 & 0.0 & 15.2\\
DI-MI-TI & 4.2 & 2.2 & 3.6 & 5.4 & 4.3 & 2.4 & 3.6 & 0.0 & 0.0 & 2.2 \\
RDI-MI-TI & 3.2 & 4.4 & 4.4 & 6.9 & 8.3 & 3.0 & 5.5 & 0.0 & 0.0 & 1.9 \\\rowcolor{Gray}
ODI-MI-TI & \textbf{15.7} & \textbf{14.7} & \textbf{17.4} & \textbf{30.4} & \textbf{32.1} & \textbf{14.1} & \textbf{26.9} & \textbf{0.3 }& \textbf{0.6}& 5.5\\
            \bottomrule%[0.15em]
\end{tabular}%
}
\caption{Targeted attack success rates (\%) against nine black-box target models with the four source models. For each attack, we also reported the average computation time to generate an adversarial example. We added experimental results of MI-TI and ATTA-MI-TI.}
\label{tab:table1}
\end{table*}
\begin{table*}[t]
\centering

\resizebox{\textwidth}{!}{%
\begin{tabular}{lccccccccc}
\toprule[0.15em]
\textbf{Source : VGG-16} & \multicolumn{9}{c}{Target model} \\
\cmidrule(l){2-10}
Object & RN-18 & RN-50 & DN-121 & Inc-v3 & Inc-v4 & Mob-v2 & IR-v2 & Adv-Inc-v3 & \begin{tabular}[c]{@{}c@{}}Ens-adv-\\ IR-v2\end{tabular} \\ \midrule
\textit{One ball} & 21.0 & 25.2 & 37.0 & 12.4 & 11.8 & 16.1 & 5.1 & 0.0 & 0.0 \\
 \textit{Two balls} & 40.1 & 33.9 & 47.1 & 36.6 & 25.0 & 25.0 & 19.4 & 0.8 & 1.0 \\
 \textit{Three balls} & \textbf{45.5} & \textbf{37.7} & \textbf{50.6} & \textbf{36.8} & \textbf{26.9} & \textbf{29.0} & \textbf{21.6} & 0.5 & 0.5 \\
  \textit{Four balls} & 32.4 & 25.3 & 35.5 & 25.5 & 17.6 & 16.7 & 14.7 & \textbf{1.4} & \textbf{2.5} \\ \bottomrule
\end{tabular}%
}
\caption{Targeted attack success rates (\%) against nine black-box target models according to the source objects. We consider the different numbers of ball objects.}
\label{tab:table2}
\end{table*}

\clearpage

\newcommand{\figw}{0.32}
\begin{figure*}[t]
     \centering
         \includegraphics[width=\figw\textwidth,trim={0.3cm 0.3cm 0.3cm 0.3cm},clip]{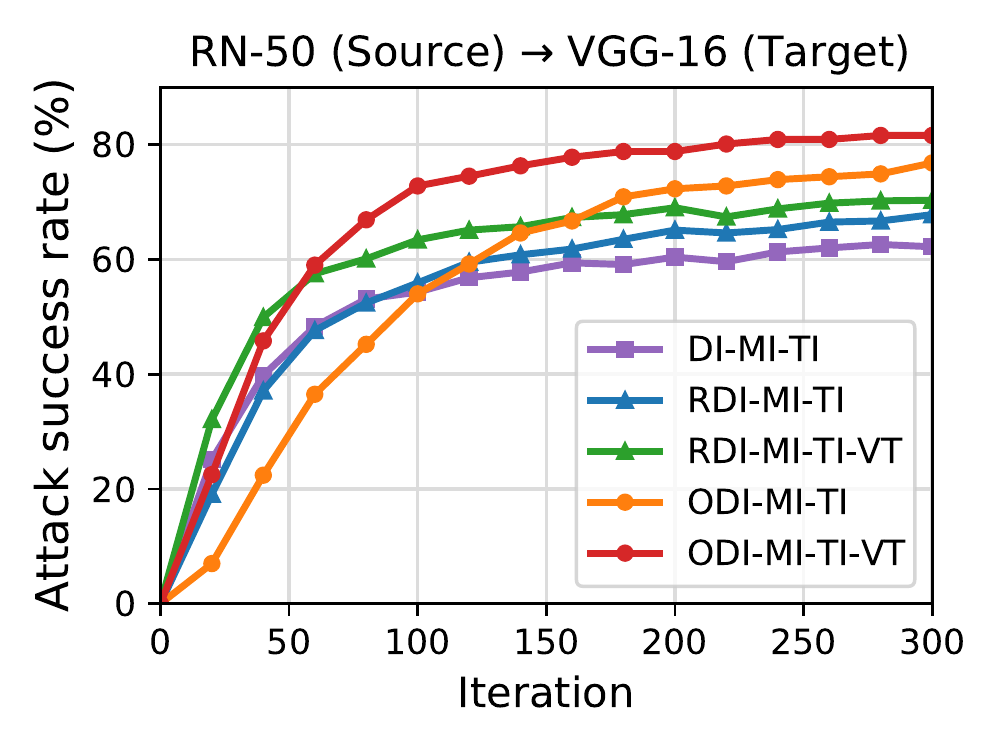}
          \includegraphics[width=\figw\textwidth,trim={0.3cm 0.3cm 0.3cm 0.3cm},clip]{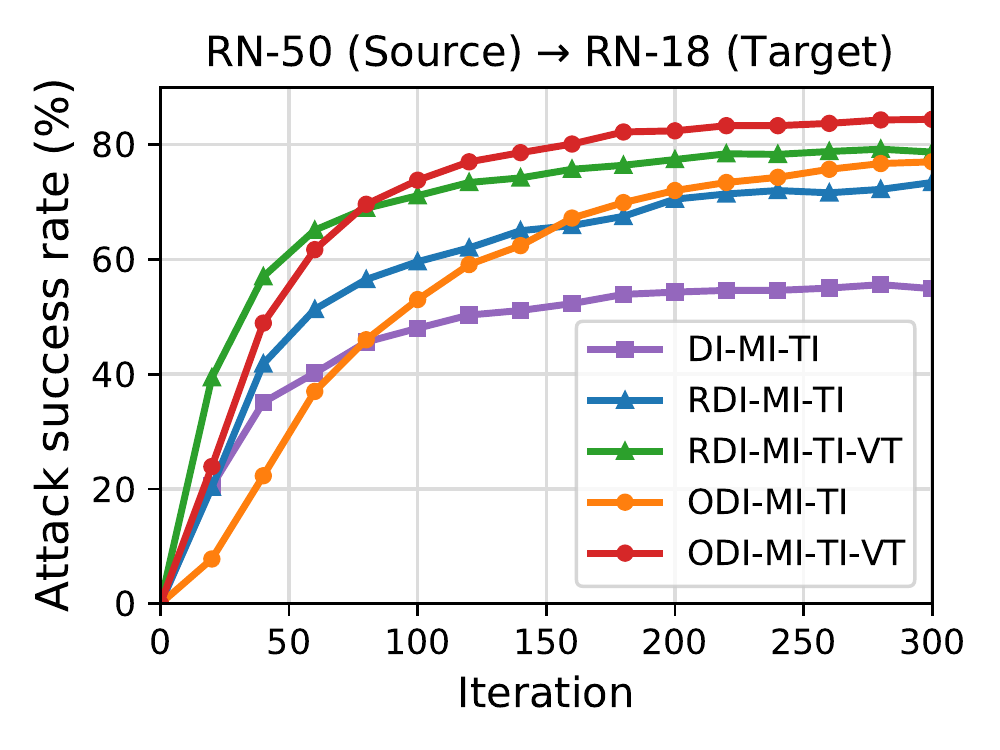}  
          \includegraphics[width=\figw\textwidth,trim={0.3cm 0.3cm 0.3cm 0.3cm},clip]{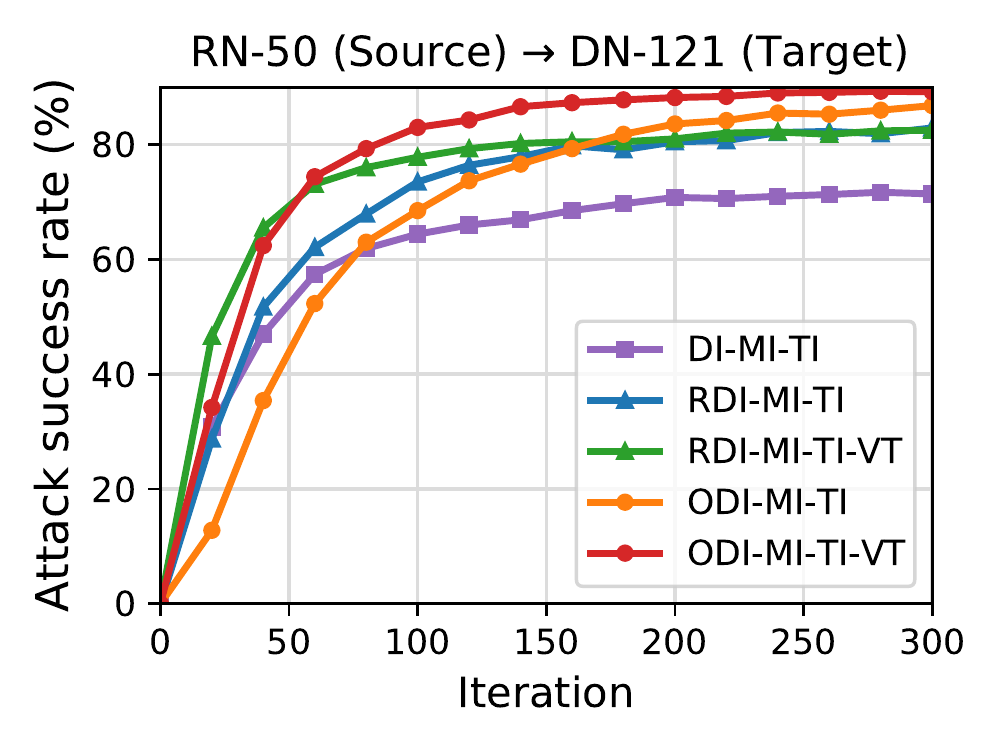}   
                   \includegraphics[width=\figw\textwidth,trim={0.3cm 0.3cm 0.3cm 0.3cm},clip]{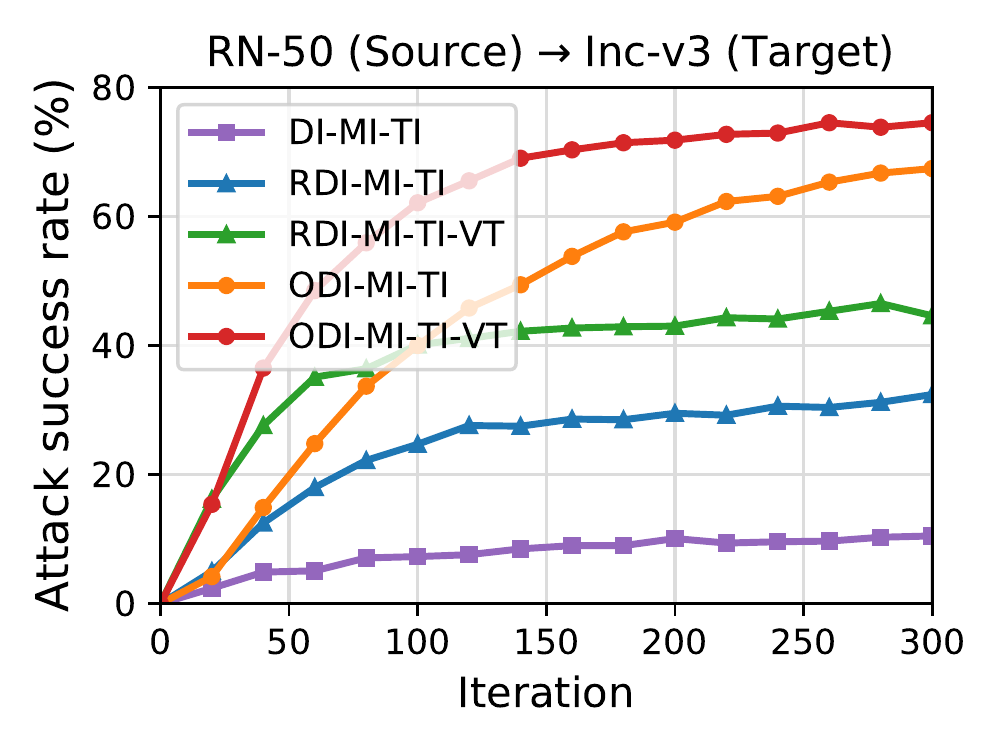}
          \includegraphics[width=\figw\textwidth,trim={0.3cm 0.3cm 0.3cm 0.3cm},clip]{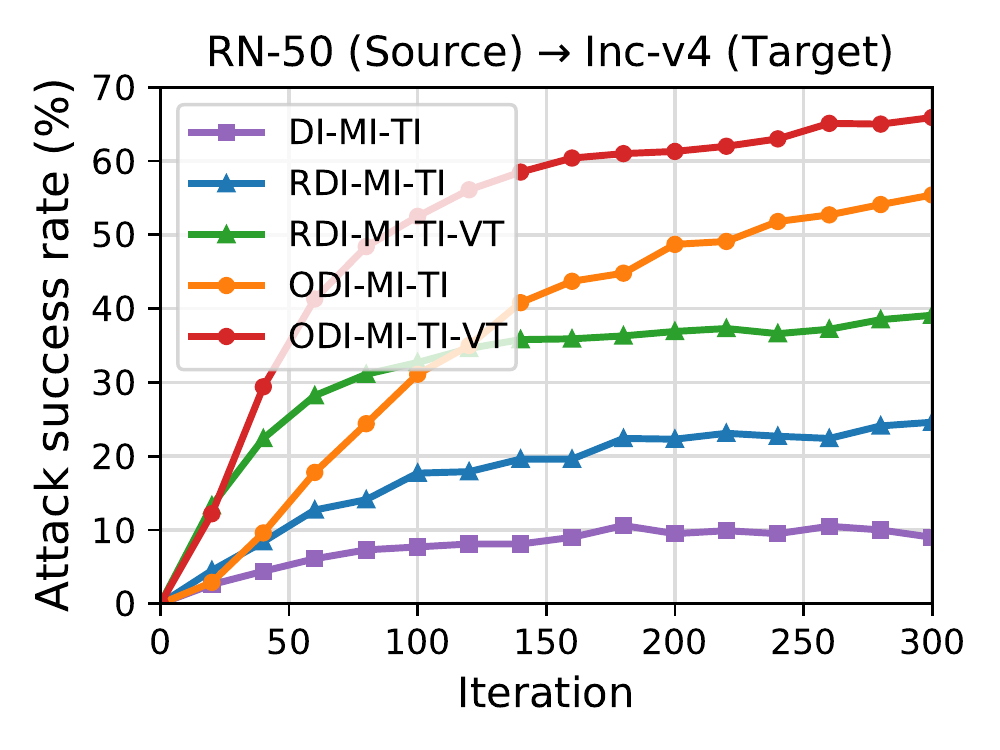}         \includegraphics[width=\figw\textwidth,trim={0.3cm 0.3cm 0.3cm 0.3cm},clip]{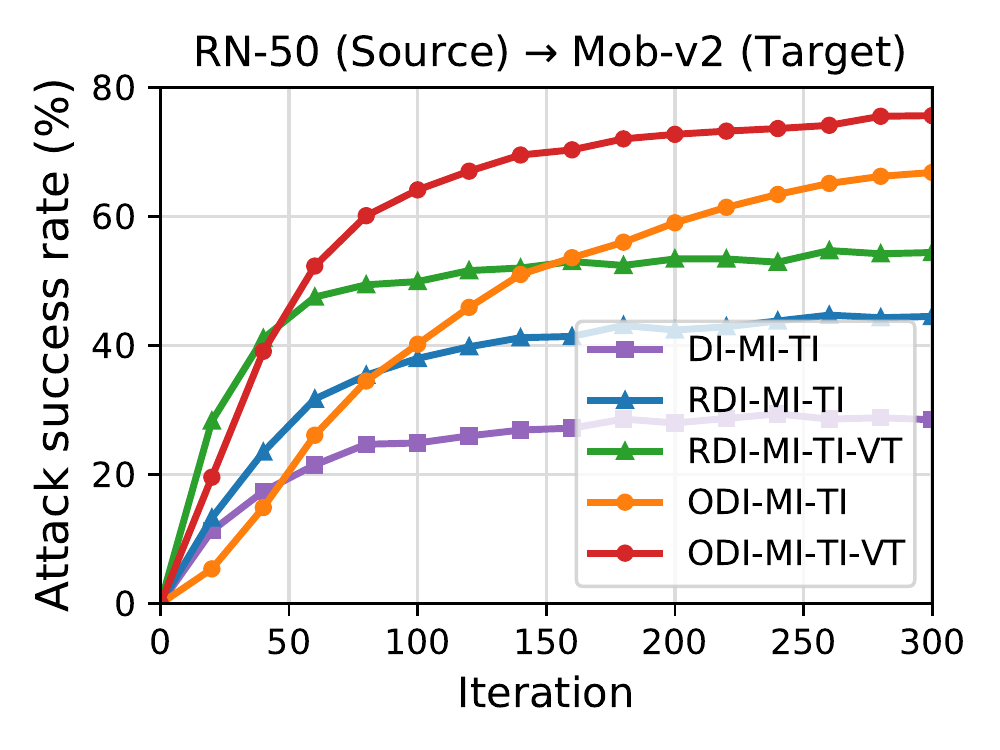}       
          \includegraphics[width=\figw\textwidth,trim={0.3cm 0.3cm 0.3cm 0.3cm},clip]{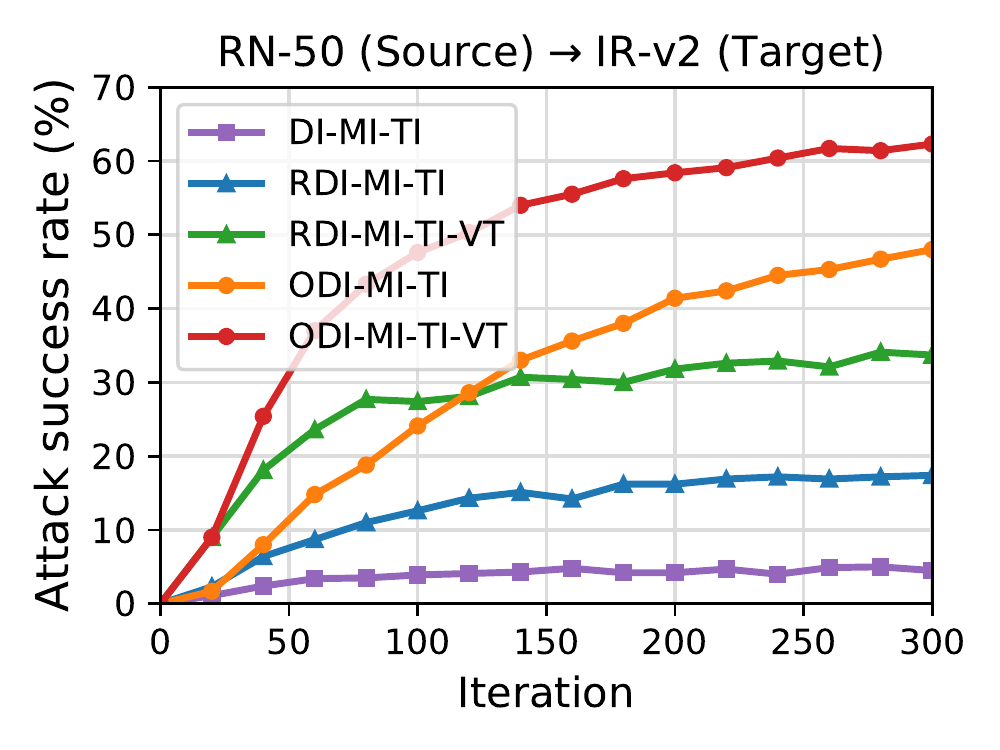}   
                   \includegraphics[width=\figw\textwidth,trim={0.3cm 0.3cm 0.3cm 0.3cm},clip]{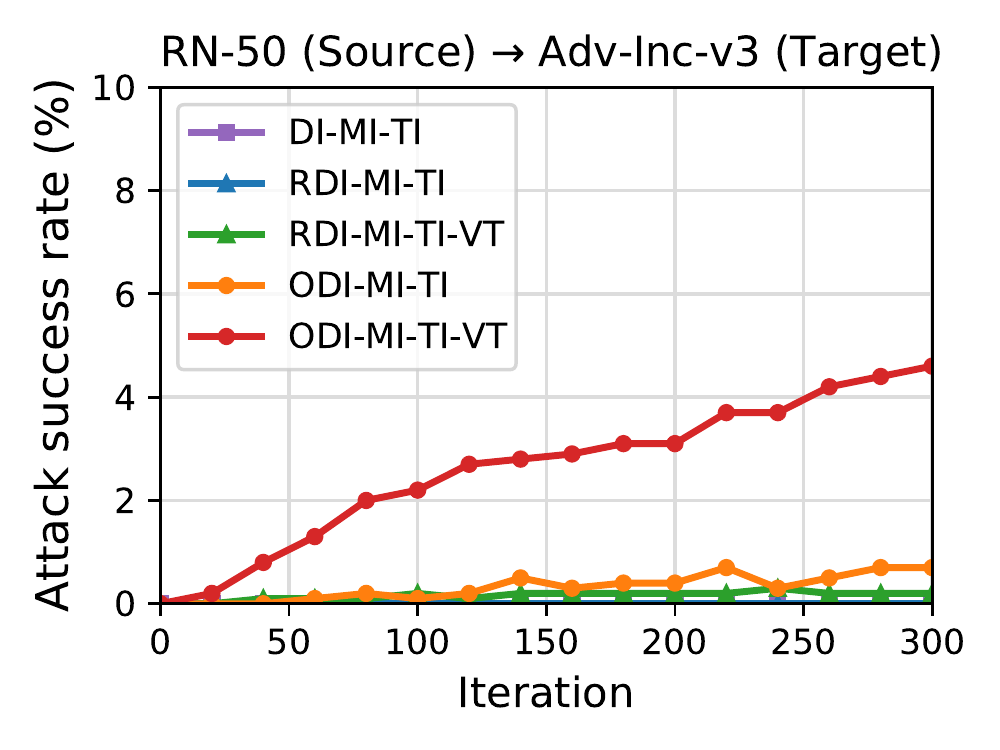}
          \includegraphics[width=\figw\textwidth,trim={0.3cm 0.3cm 0.3cm 0.3cm},clip]{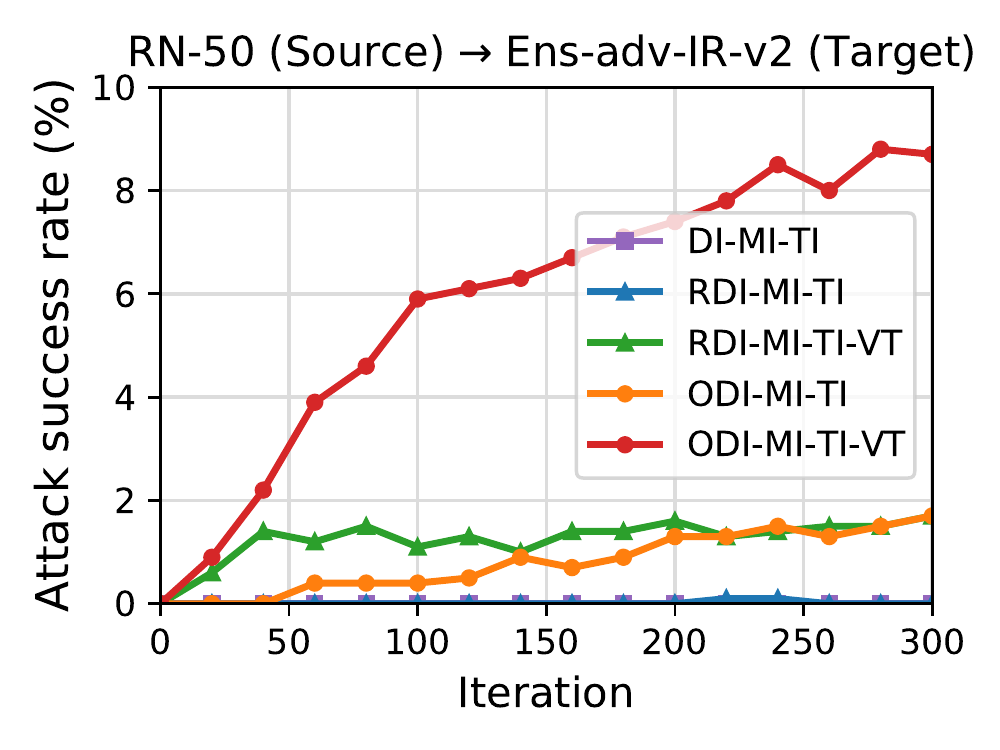}
        \caption{Targeted attack success rates (\%) according to the number of iterations. The source model is RN-50.}
        \label{fig:fig1}
\end{figure*}
\begin{figure*}[t]
     \centering
         \includegraphics[width=\figw\textwidth,trim={0.3cm 0.3cm 0.3cm 0.3cm},clip]{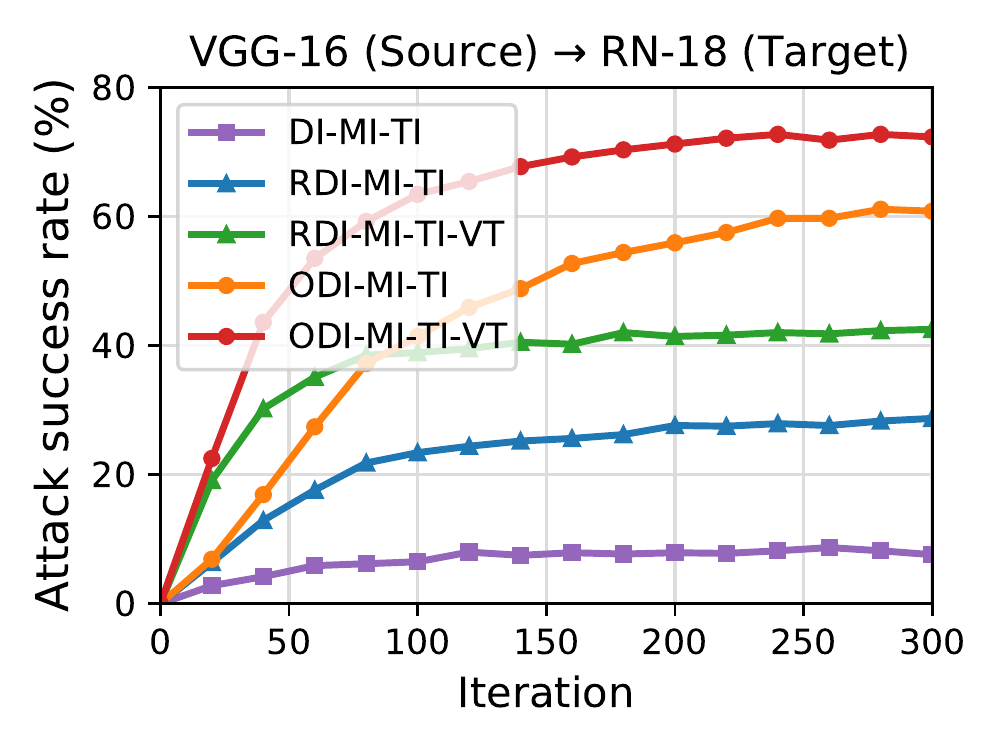}
          \includegraphics[width=\figw\textwidth,trim={0.3cm 0.3cm 0.3cm 0.3cm},clip]{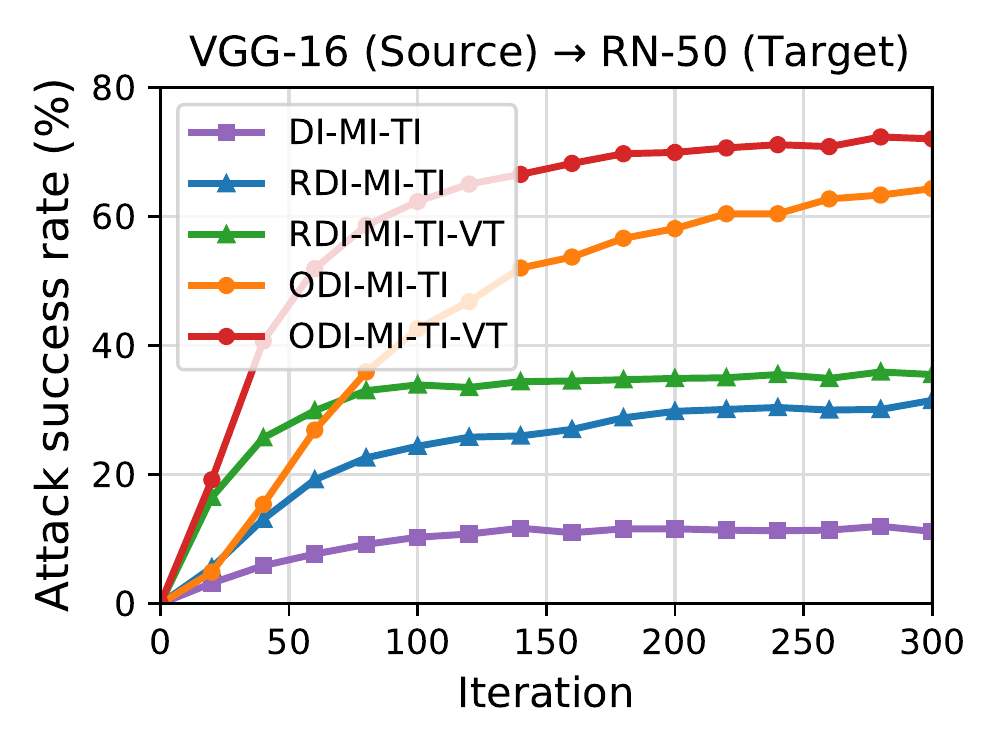}  
          \includegraphics[width=\figw\textwidth,trim={0.3cm 0.3cm 0.3cm 0.3cm},clip]{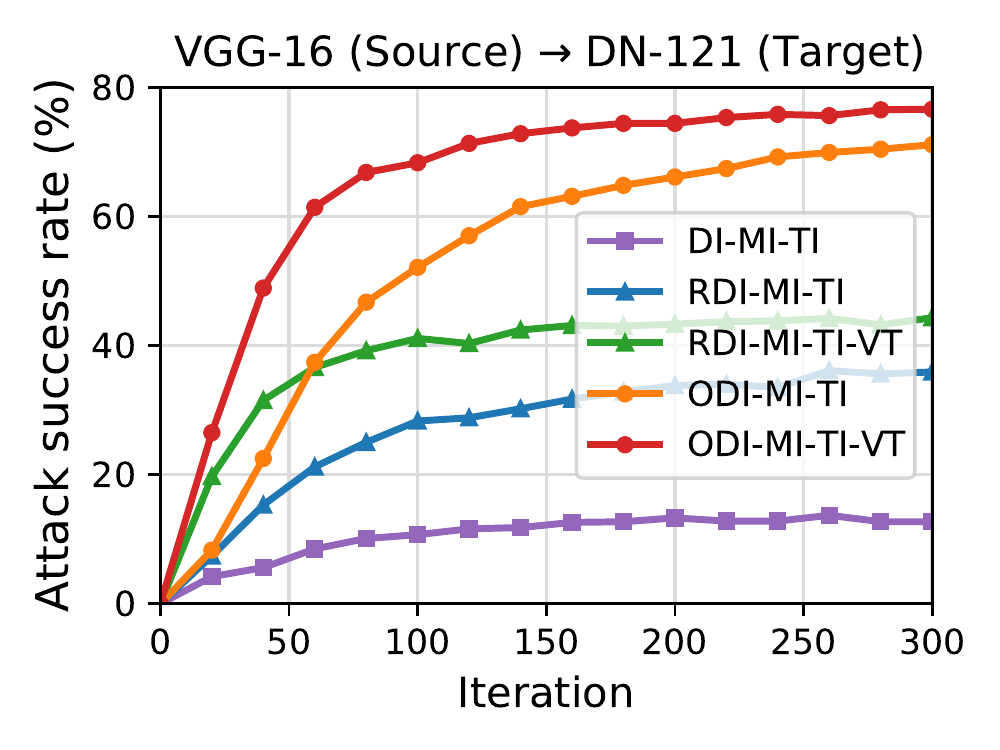}   
                   \includegraphics[width=\figw\textwidth,trim={0.3cm 0.3cm 0.3cm 0.3cm},clip]{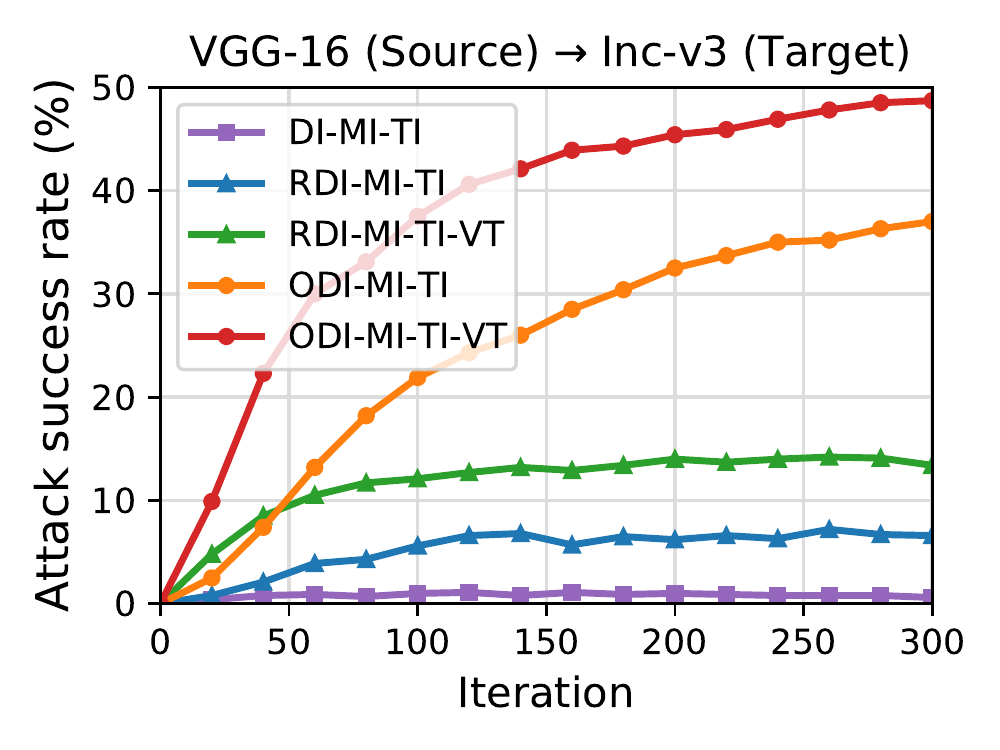}
          \includegraphics[width=\figw\textwidth,trim={0.3cm 0.3cm 0.3cm 0.3cm},clip]{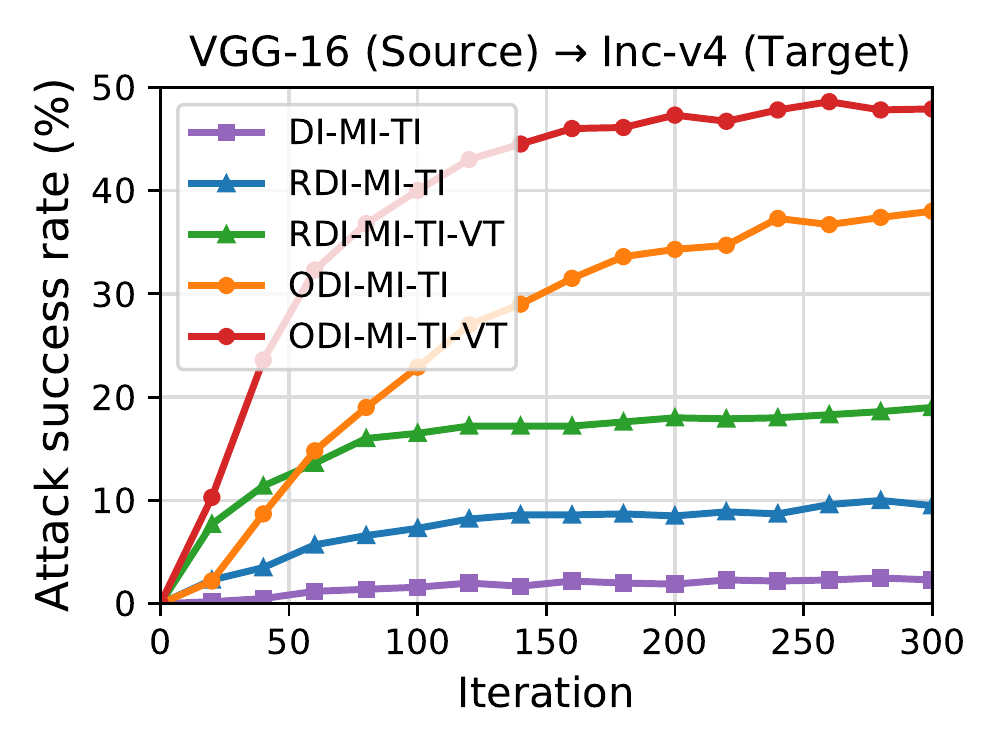}         \includegraphics[width=\figw\textwidth,trim={0.3cm 0.3cm 0.3cm 0.3cm},clip]{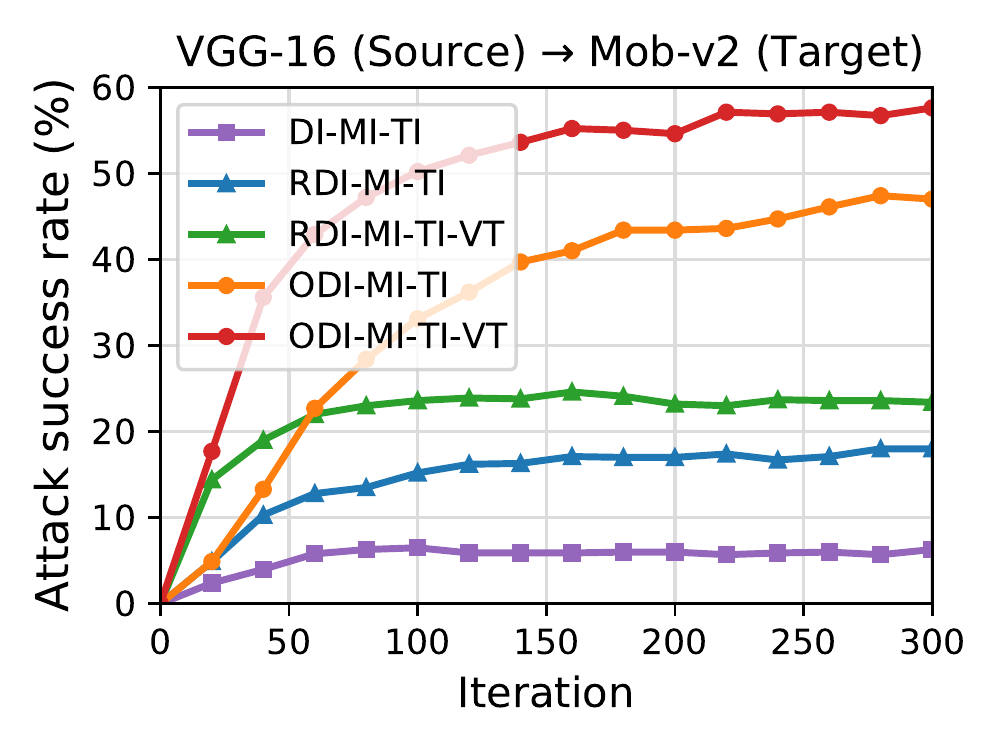}       
          \includegraphics[width=\figw\textwidth,trim={0.3cm 0.3cm 0.3cm 0.3cm},clip]{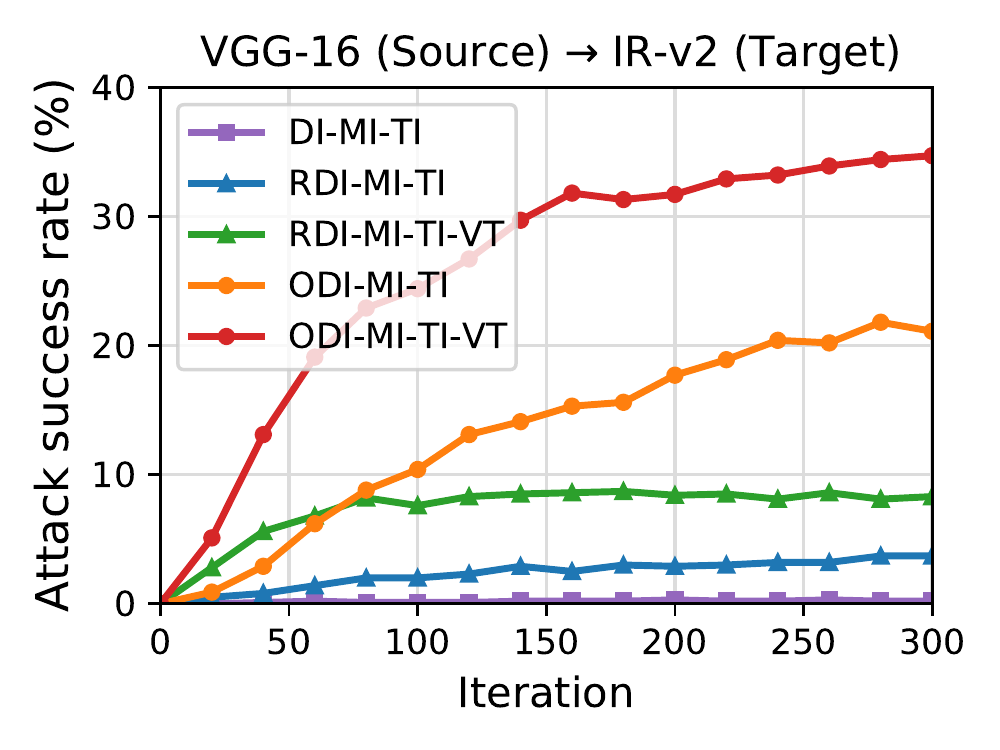}   
                   \includegraphics[width=\figw\textwidth,trim={0.3cm 0.3cm 0.3cm 0.3cm},clip]{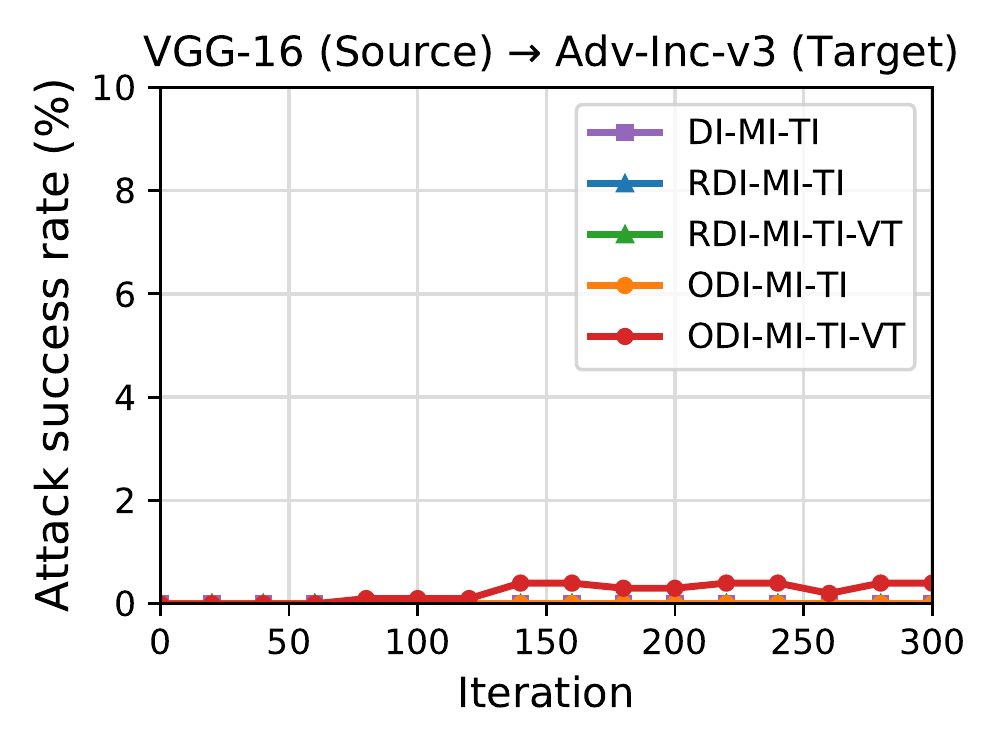}
          \includegraphics[width=\figw\textwidth,trim={0.3cm 0.3cm 0.3cm 0.3cm},clip]{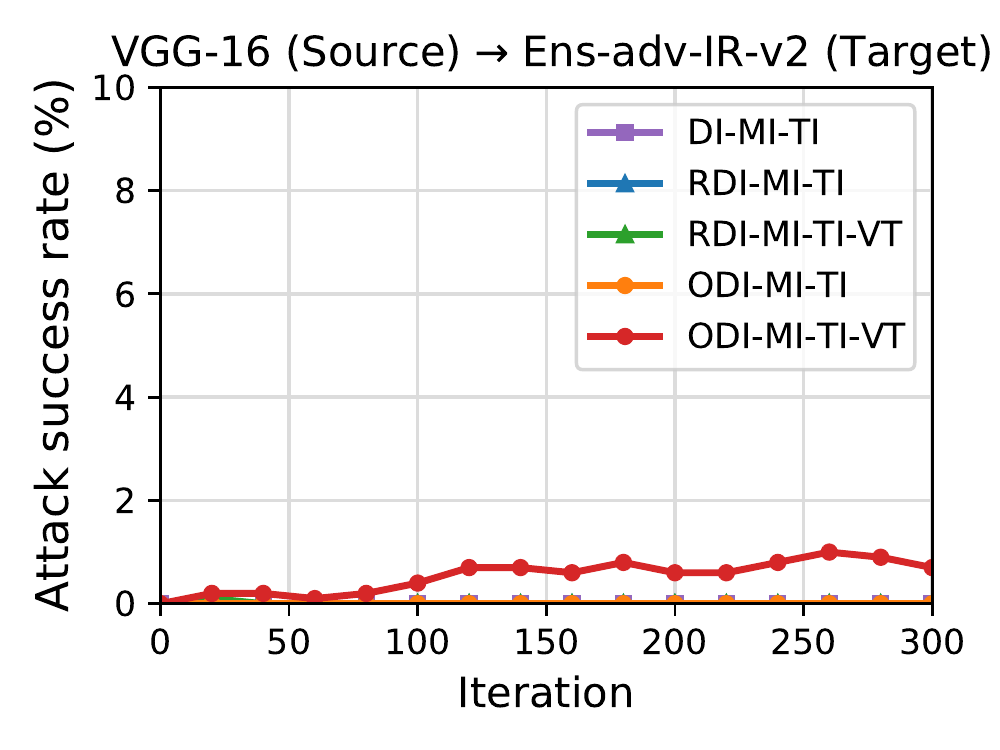}
        \caption{Targeted attack success rates (\%) according to the number of iterations. The source model is VGG-16.}
        \label{fig:fig2}
\end{figure*}

\begin{figure*}[t]
     \centering
         \includegraphics[width=\figw\textwidth,trim={0.3cm 0.3cm 0.3cm 0.3cm},clip]{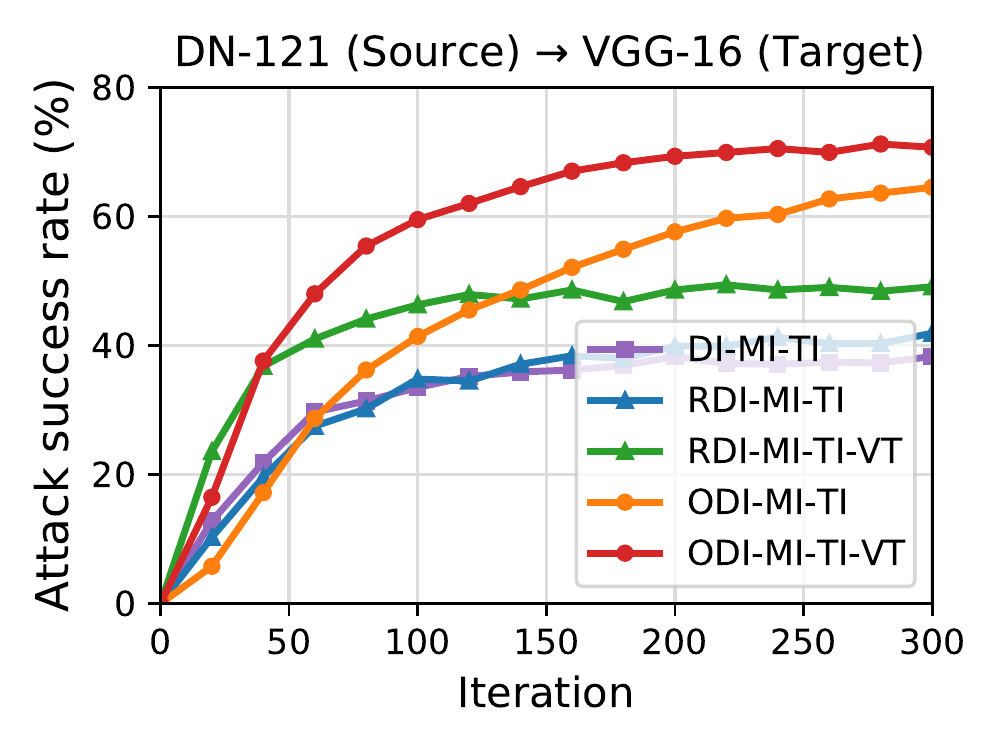}
          \includegraphics[width=\figw\textwidth,trim={0.3cm 0.3cm 0.3cm 0.3cm},clip]{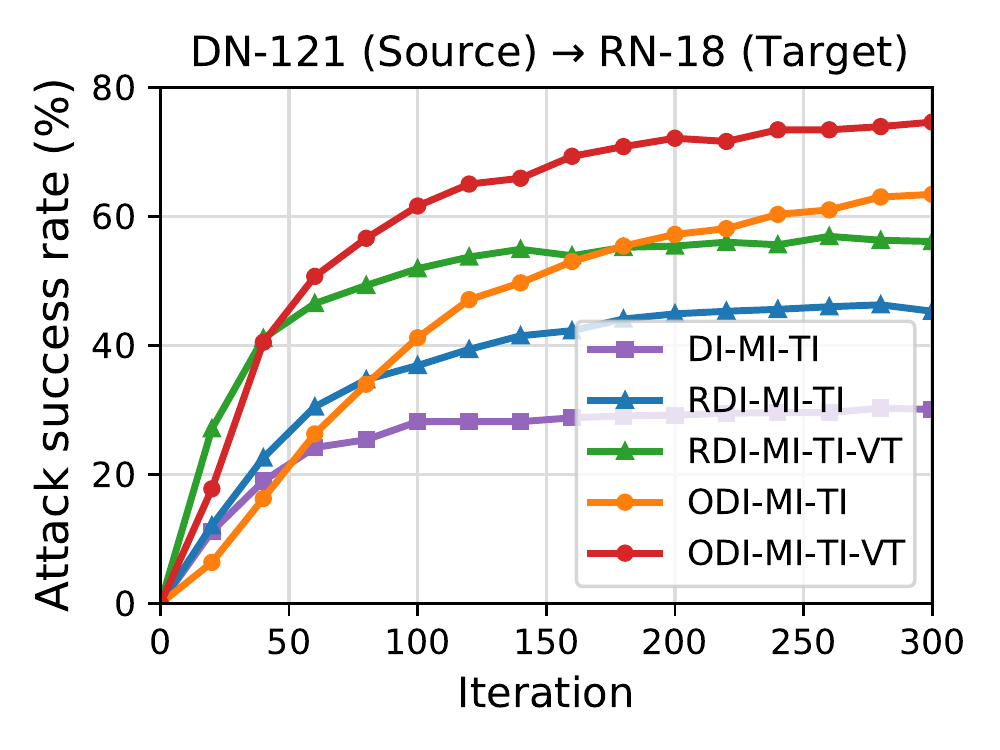}  
          \includegraphics[width=\figw\textwidth,trim={0.3cm 0.3cm 0.3cm 0.3cm},clip]{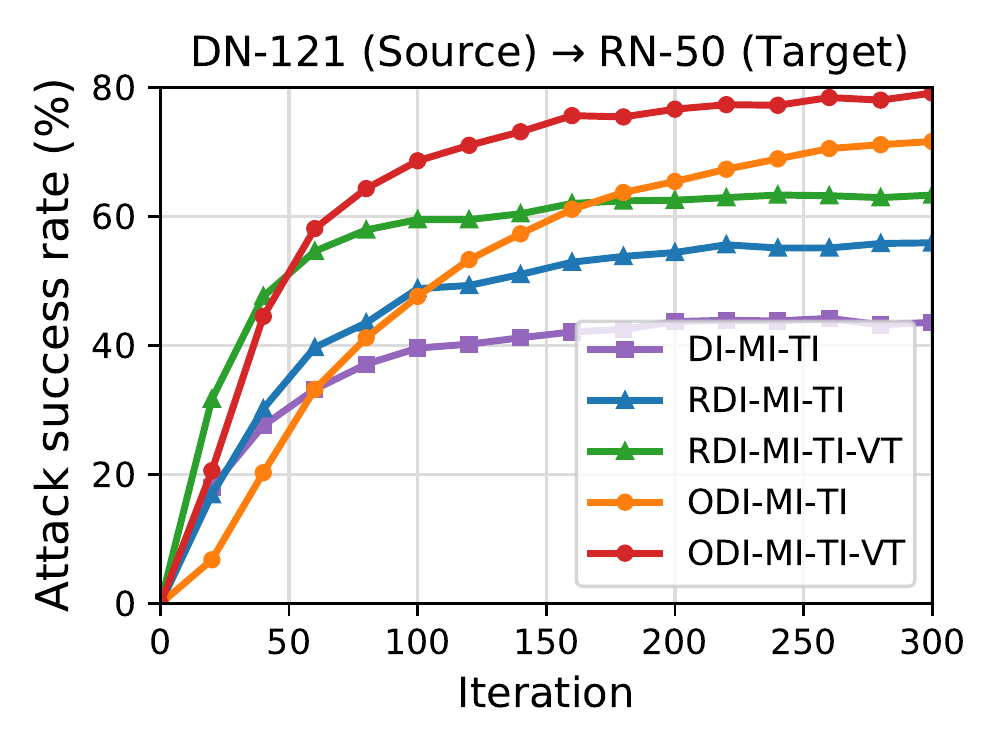}   
                   \includegraphics[width=\figw\textwidth,trim={0.3cm 0.3cm 0.3cm 0.3cm},clip]{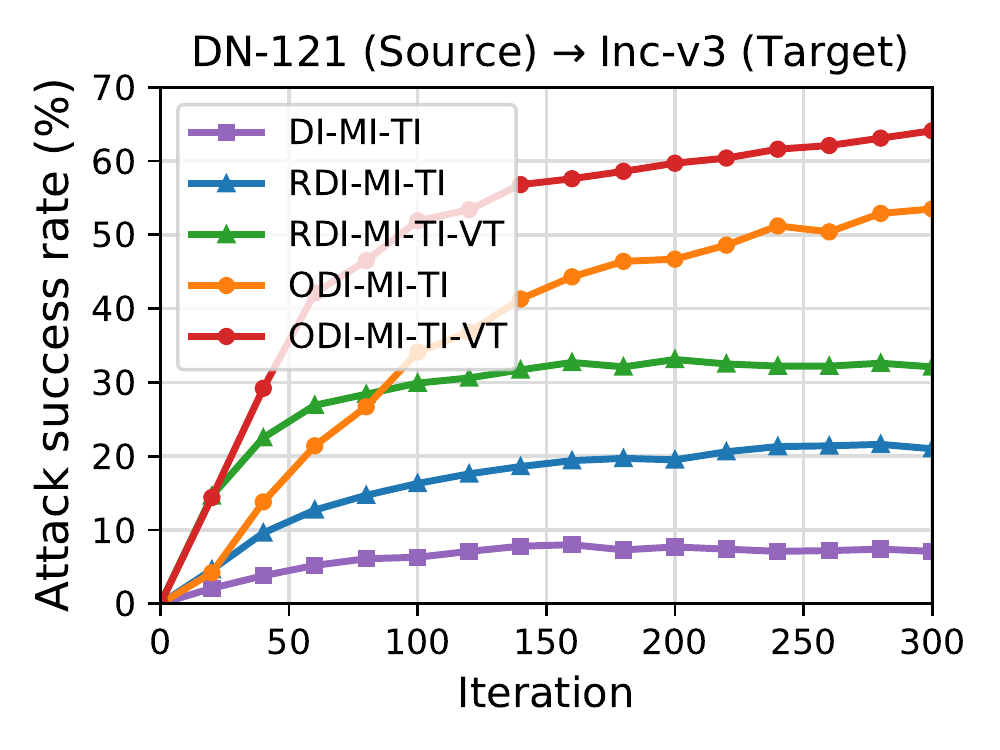}
          \includegraphics[width=\figw\textwidth,trim={0.3cm 0.3cm 0.3cm 0.3cm},clip]{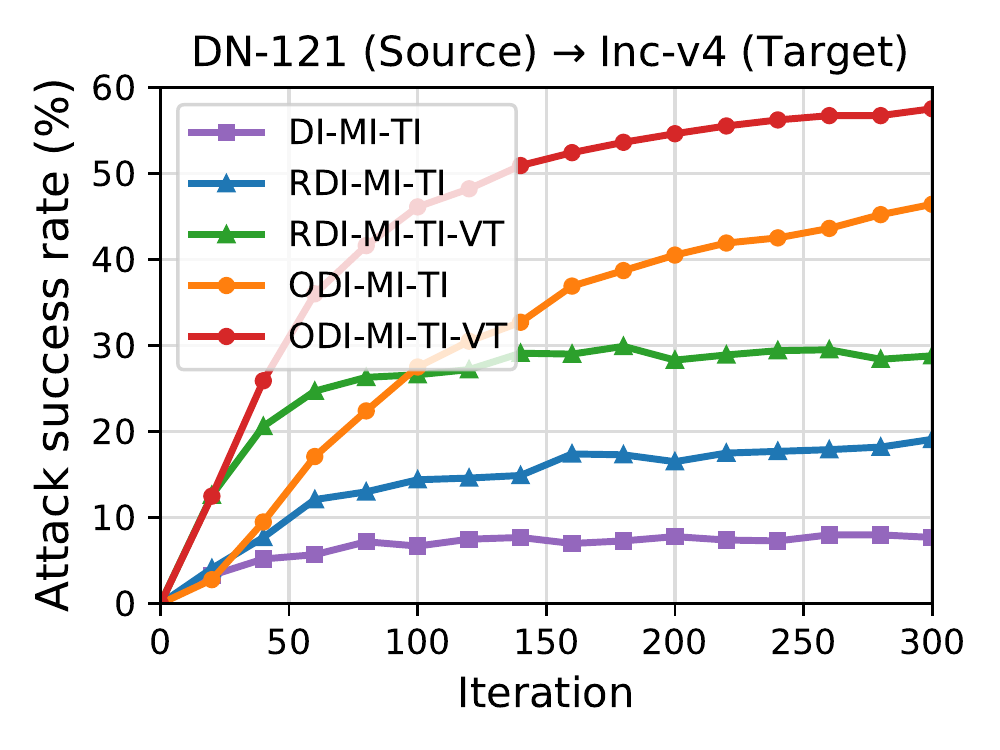}         \includegraphics[width=\figw\textwidth,trim={0.3cm 0.3cm 0.3cm 0.3cm},clip]{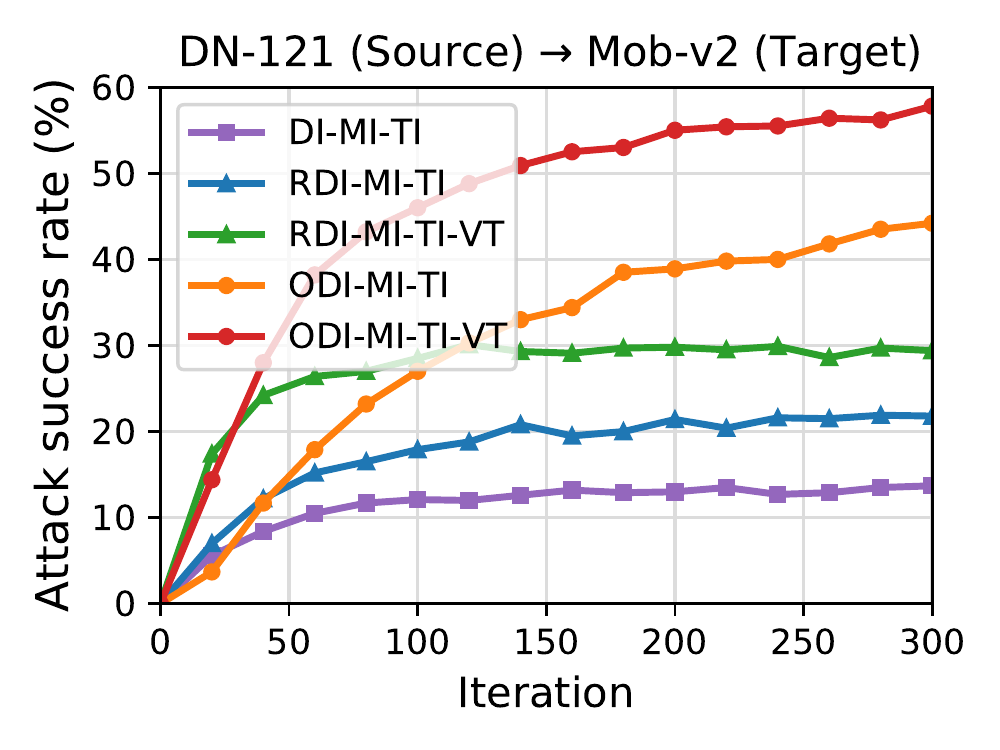}       
          \includegraphics[width=\figw\textwidth,trim={0.3cm 0.3cm 0.3cm 0.3cm},clip]{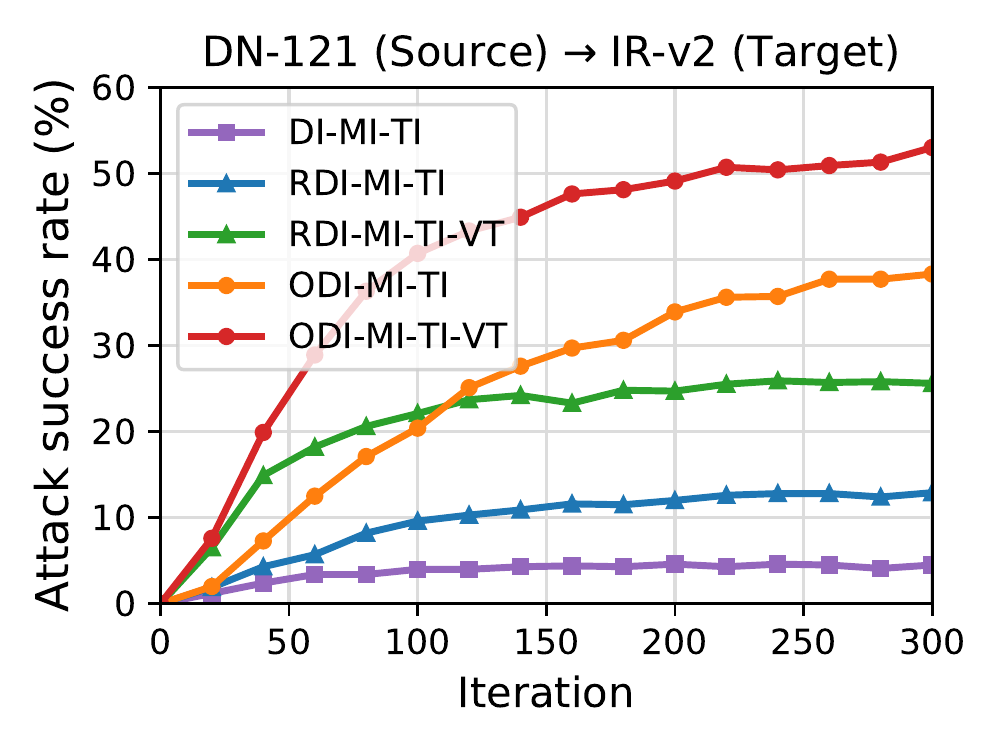}   
                   \includegraphics[width=\figw\textwidth,trim={0.3cm 0.3cm 0.3cm 0.3cm},clip]{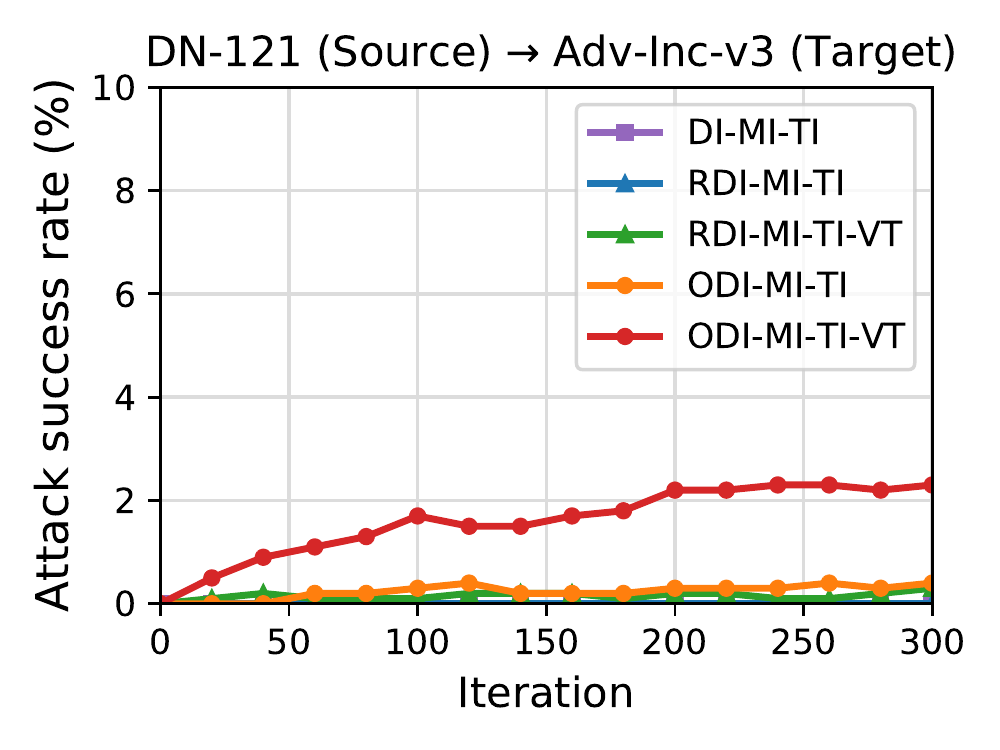}
          \includegraphics[width=\figw\textwidth,trim={0.3cm 0.3cm 0.3cm 0.3cm},clip]{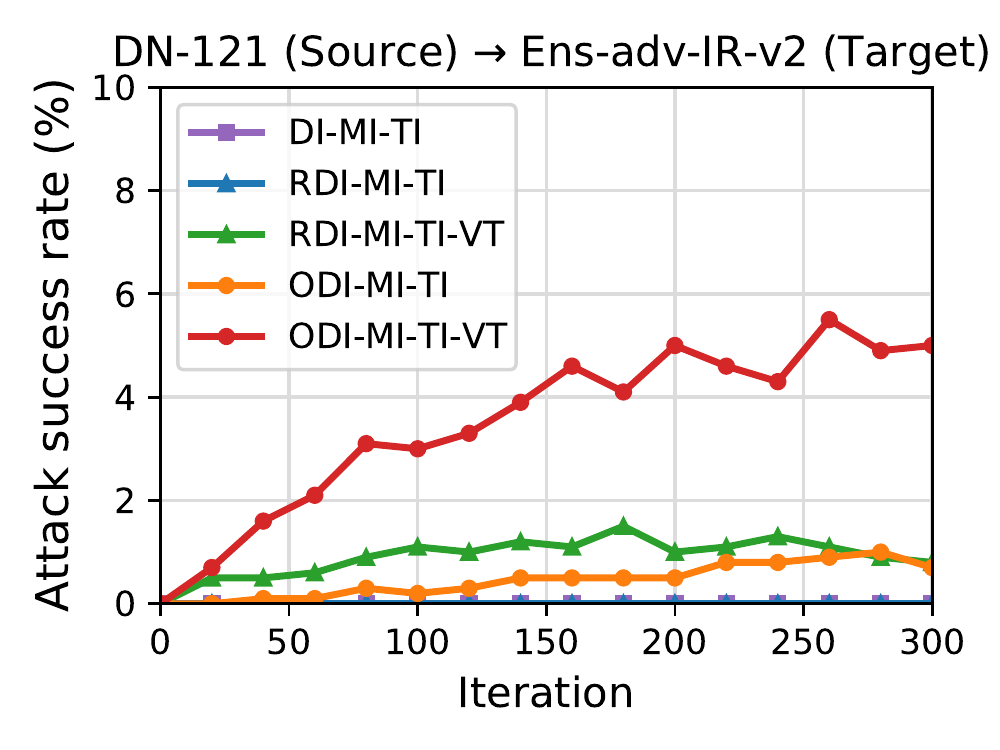}
        \caption{Targeted attack success rates (\%) according to the number of iterations. The source model is DN-121.}
        \label{fig:fig3}
\end{figure*}

\begin{figure*}[t]
     \centering
         \includegraphics[width=\figw\textwidth,trim={0.3cm 0.3cm 0.3cm 0.3cm},clip]{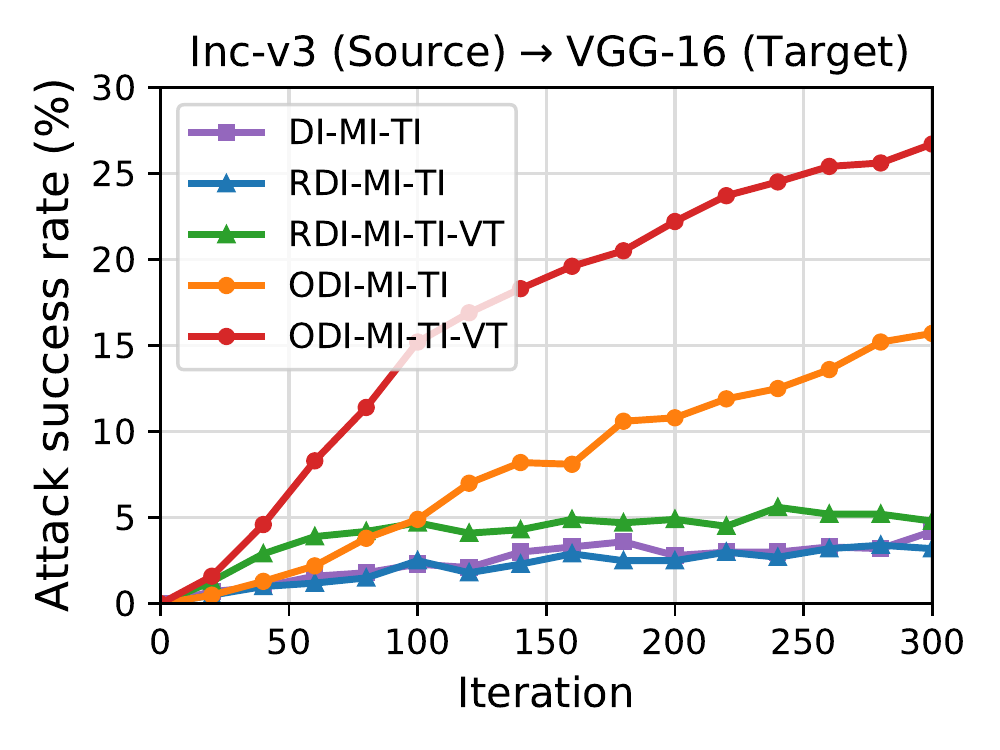}
          \includegraphics[width=\figw\textwidth,trim={0.3cm 0.3cm 0.3cm 0.3cm},clip]{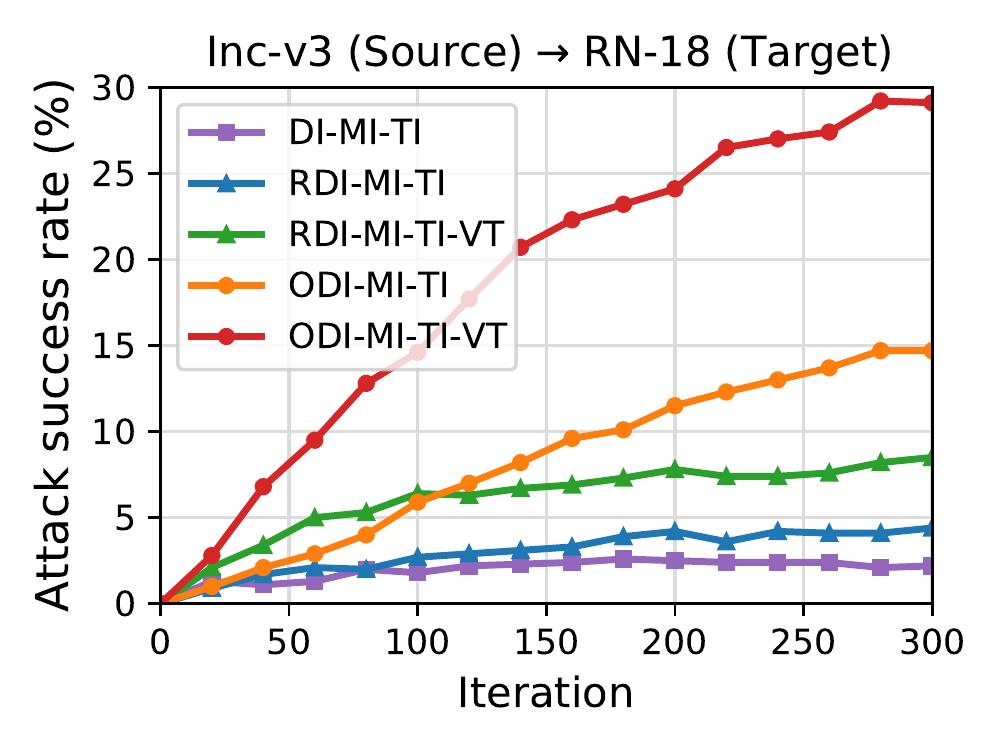}  
          \includegraphics[width=\figw\textwidth,trim={0.3cm 0.3cm 0.3cm 0.3cm},clip]{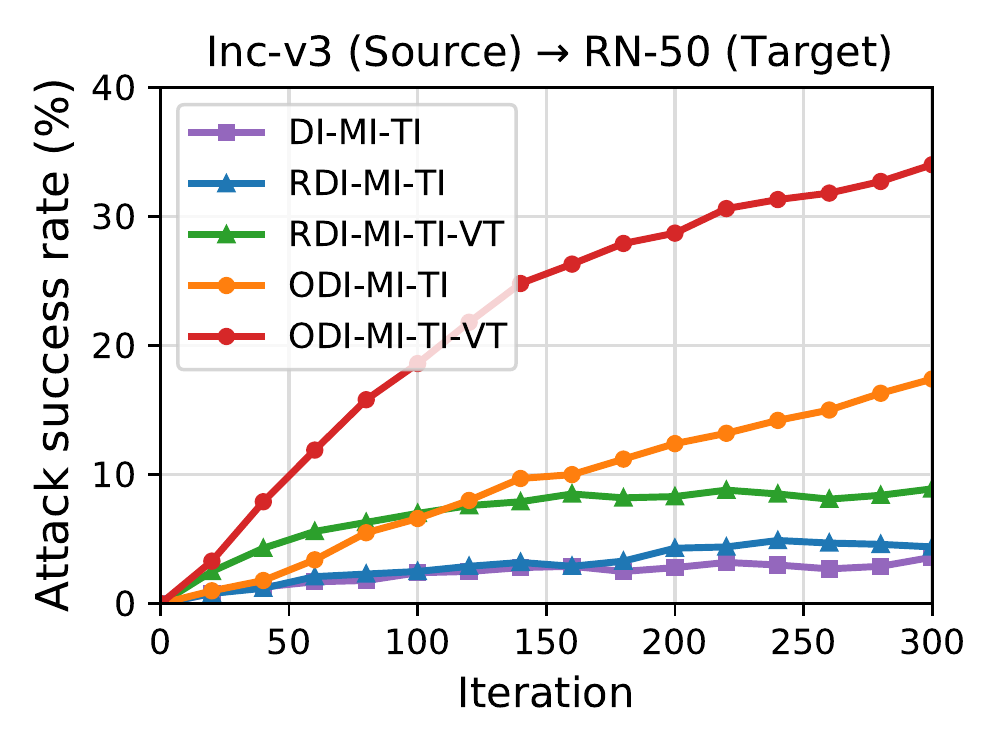}   
                   \includegraphics[width=\figw\textwidth,trim={0.3cm 0.3cm 0.3cm 0.3cm},clip]{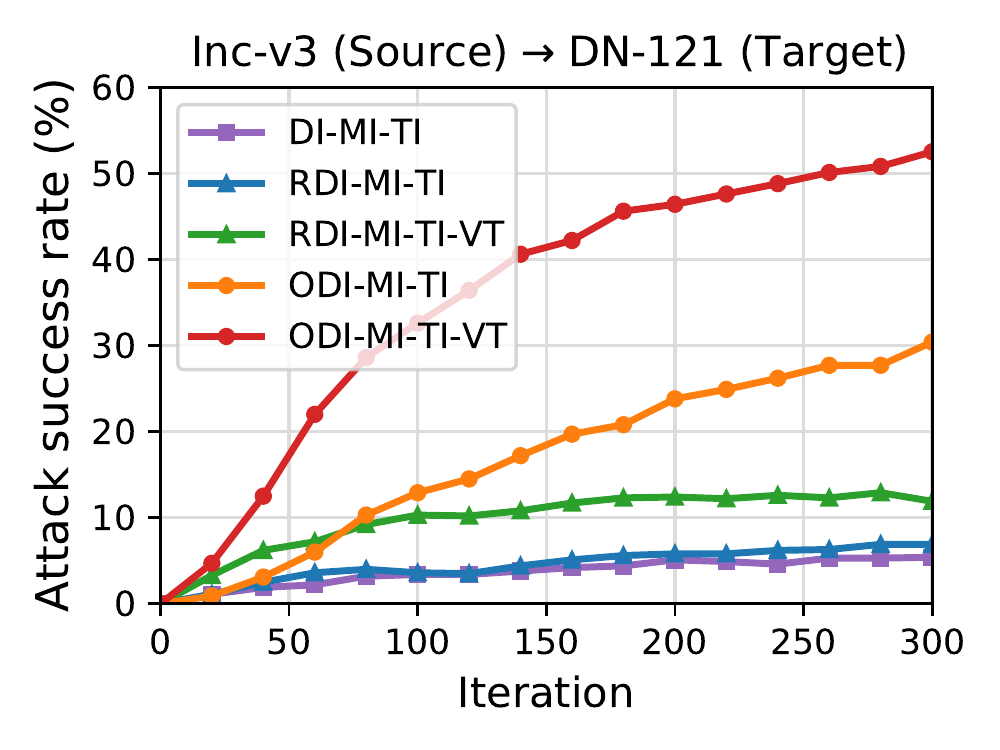}
          \includegraphics[width=\figw\textwidth,trim={0.3cm 0.3cm 0.3cm 0.3cm},clip]{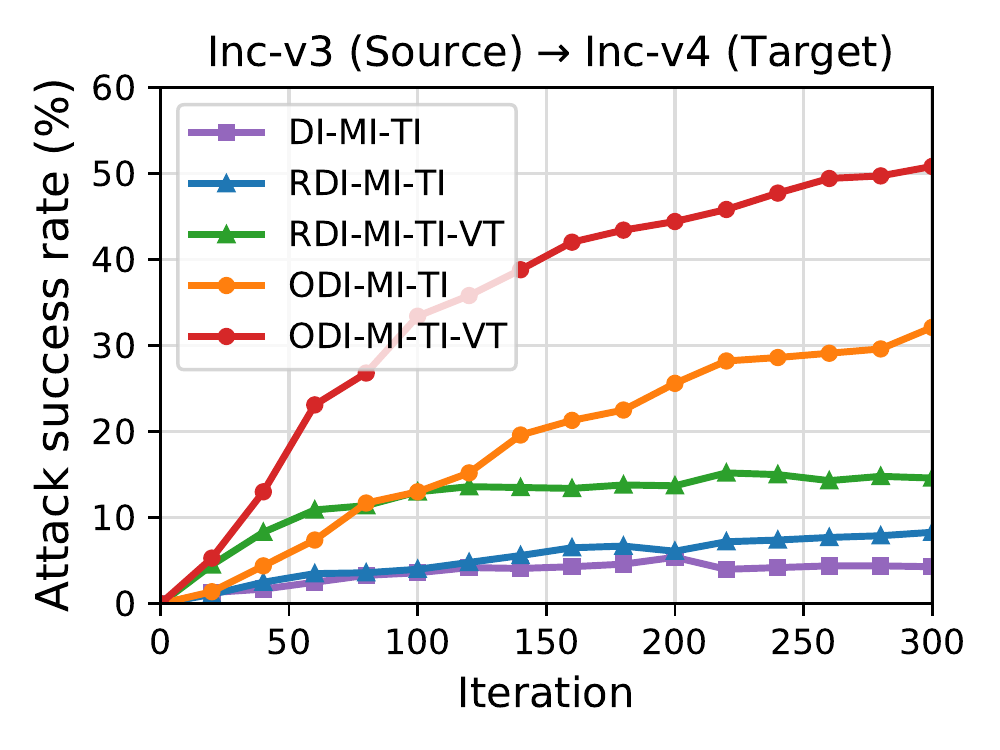}         \includegraphics[width=\figw\textwidth,trim={0.3cm 0.3cm 0.3cm 0.3cm},clip]{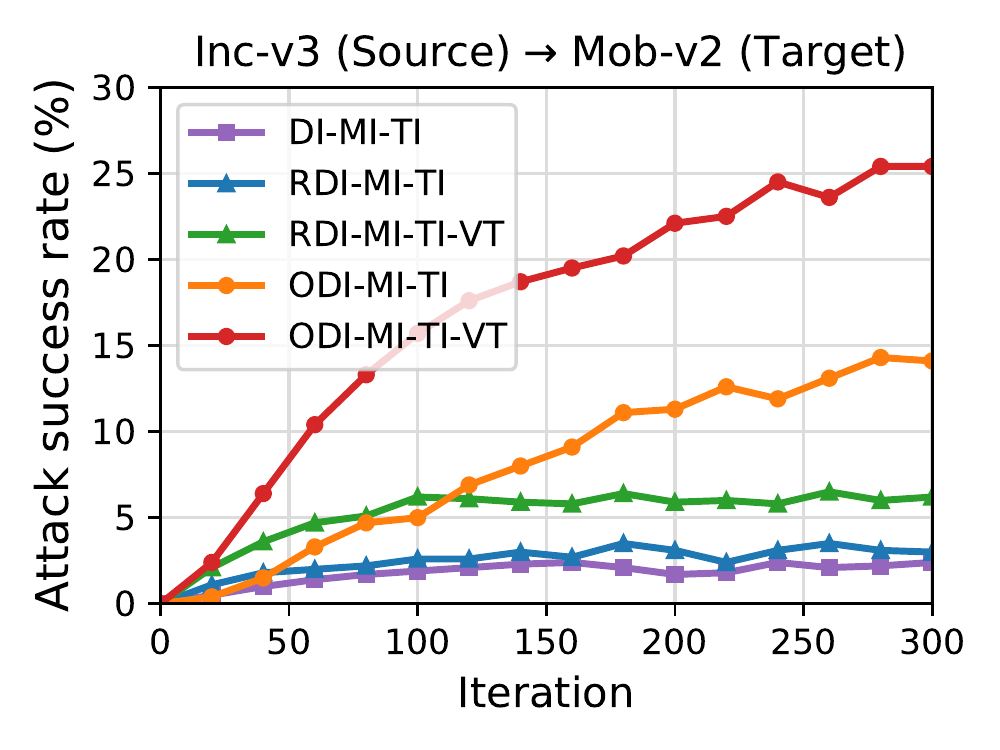}       
          \includegraphics[width=\figw\textwidth,trim={0.3cm 0.3cm 0.3cm 0.3cm},clip]{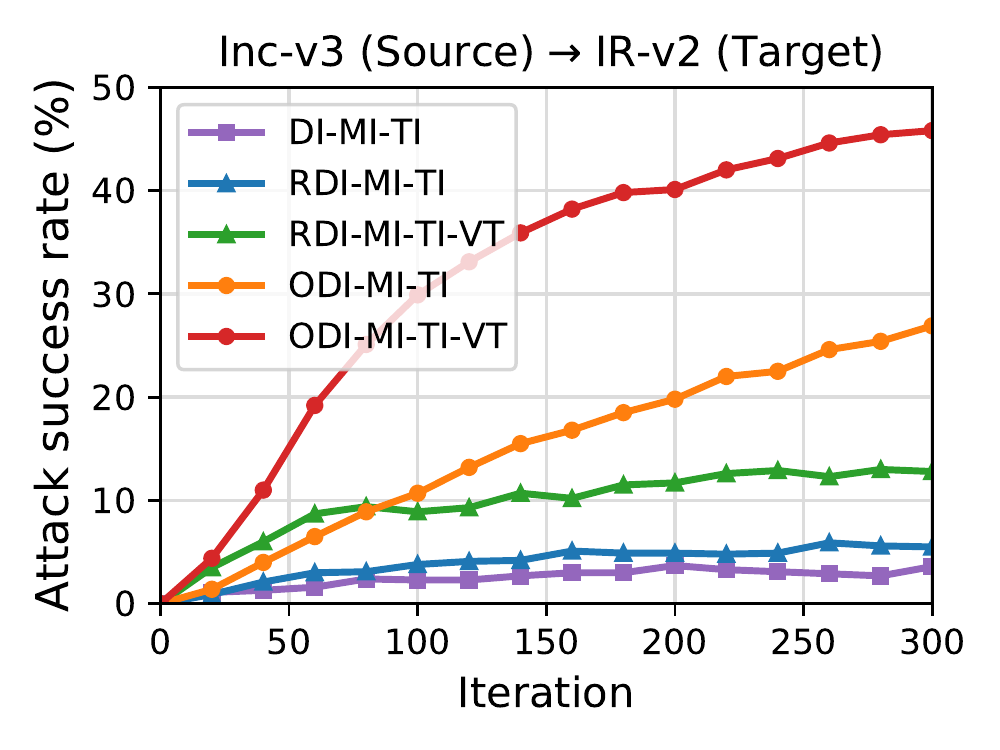}   
                   \includegraphics[width=\figw\textwidth,trim={0.3cm 0.3cm 0.3cm 0.3cm},clip]{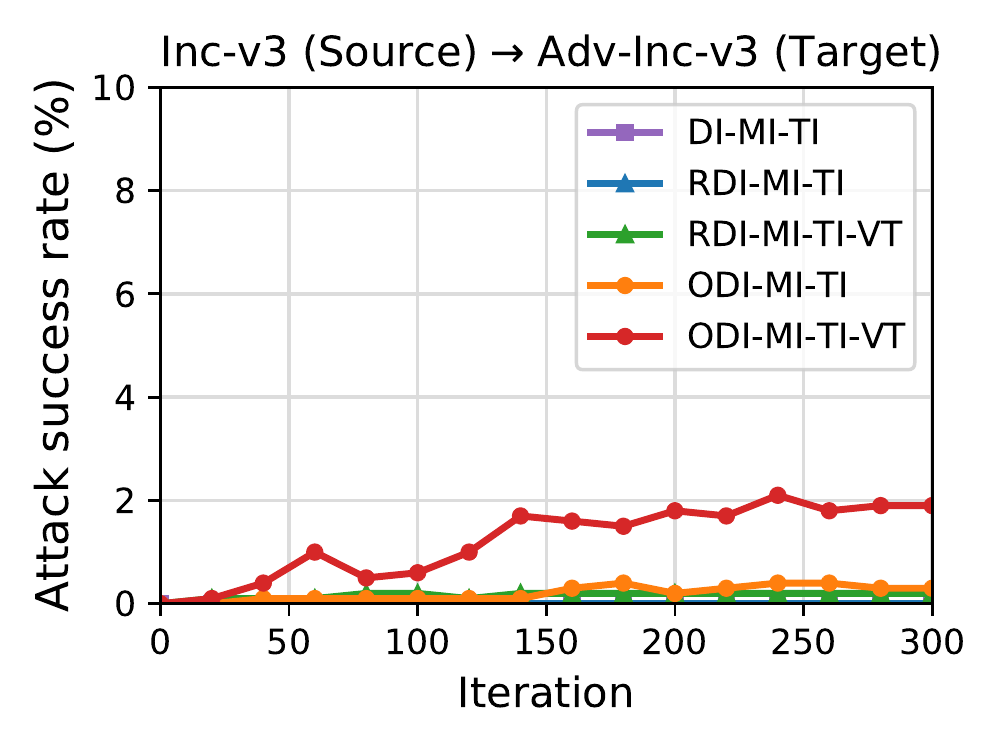}
          \includegraphics[width=\figw\textwidth,trim={0.3cm 0.3cm 0.3cm 0.3cm},clip]{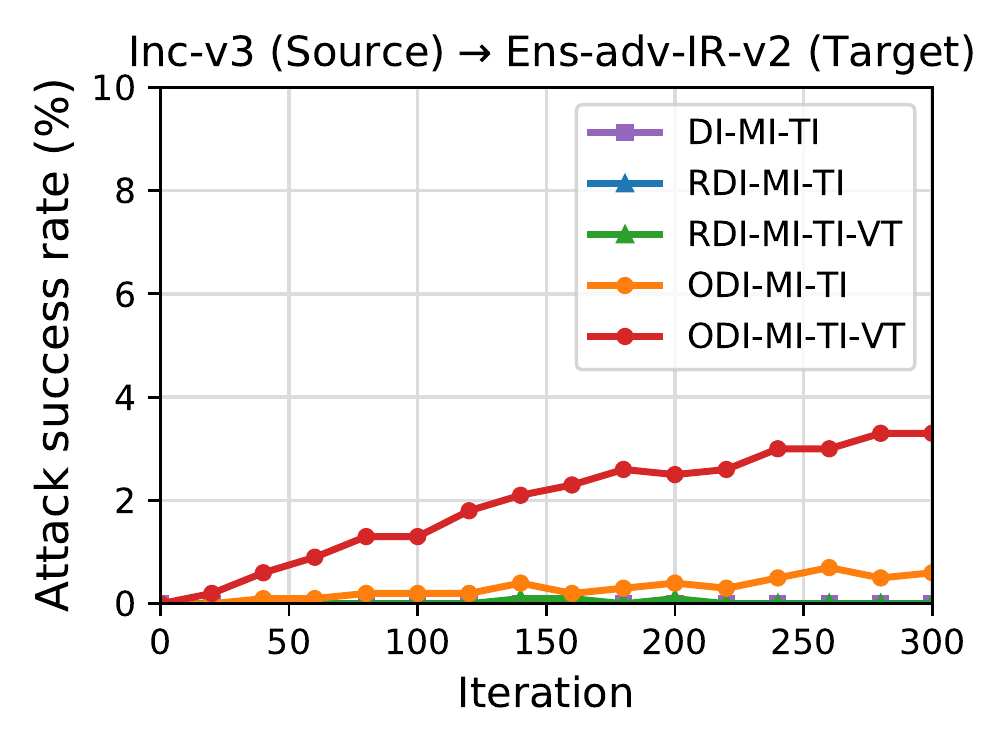}
        \caption{Targeted attack success rates (\%) according to the number of iterations. The source model is Inc-v3.}
        \label{fig:fig4}
\end{figure*}